\documentclass[11pt]{article}
\usepackage{acl}

\usepackage{times}
\usepackage{latexsym}

\usepackage{CJKutf8}
\usepackage{adjustbox}

\usepackage[T2A,T1]{fontenc}

\usepackage[utf8]{inputenc}
\usepackage[table]{colortbl}
\definecolor{olive1}{rgb}{0.88, 0.74, 0.41}
\definecolor{teal1}{rgb}{0.29, 0.74, 0.70}

\usepackage{setspace}

\usepackage{microtype}
\usepackage{hyperref}
\usepackage{url}
\definecolor{darkblue}{rgb}{0, 0, 0.5}
\hypersetup{colorlinks=true, citecolor=darkblue, linkcolor=darkblue, urlcolor=darkblue}

\usepackage{graphicx}
\usepackage{tabularx}
\usepackage{xcolor}
\usepackage{array}
\usepackage{soul}
\usepackage{multirow}
\usepackage{enumitem}%
\usepackage{amsmath}
\usepackage[capitalize]{cleveref}
\usepackage{makecell}
\usepackage{xstring}
\usepackage{color}
\usepackage{booktabs}
\usepackage{worldflags}
\usepackage{longtable}
\setlist[enumerate]{itemsep=0mm}

\setkeys{Gin}{width=\textwidth}
\makeatletter
\g@addto@macro\@floatboxreset\centering
\makeatother

\makeatletter
\DeclareRobustCommand\onedot{\futurelet\@let@token\@onedot}
\def\@onedot{\ifx\@let@token.\else.\null\fi\xspace}

\makeatother

\newcommand{\specialurl}[1]{\texttt{\url{#1}}}

\title{MAiDE-up: Multilingual Deception Detection of \\ GPT-generated Hotel Reviews}

 \author{Oana Ignat \hspace{5pt} 
 Xiaomeng Xu \hspace{5pt}  
Rada Mihalcea \\
University of Michigan - Ann Arbor, USA  \\
\textit{\{oignat, phoebexu, mihalcea\}@umich.edu} \\  }

\begin{document}

\maketitle

\begin{abstract}
Deceptive reviews are becoming increasingly common, especially given the increase in performance and the prevalence of LLMs. While work to date has addressed the development of models to differentiate between truthful and deceptive human reviews, much less is known about the distinction between real reviews and AI-authored fake reviews. Moreover, most of the research so far has focused primarily on English, with very little work dedicated to other languages. In this paper, we compile and make publicly available the \textsc{MAiDE-up} dataset, consisting of 10,000 real and 10,000 AI-generated fake hotel reviews, balanced across ten languages. 
Using this dataset, we conduct extensive linguistic analyses to (1) compare the AI fake hotel reviews to real hotel reviews, and (2) identify the factors that influence the deception detection model performance.
We explore the effectiveness of several models for deception detection in hotel reviews across three main dimensions: sentiment, location, and language. 
We find that these dimensions influence how well we can detect AI-generated fake reviews.
Our dataset can be accessed alongside our generation and classification models at
\url{https://github.com/MichiganNLP/multilingual_reviews_deception}.
\end{abstract}

\section{Introduction}
Recent advancements in Natural Language Generation (NLG) technology have greatly improved the quality of LLM-generated text. One notable example is OpenAI's ChatGPT, which has demonstrated exceptional performance in tasks such as story generation~\citep{yuan2022wordcraft}, question answering~\citep{bahak2023evaluating}, essay writing~\citep{stokel2022ai}, and coding~\citep{becker2023programming}. However, this newfound ability to produce highly efficient, human-like texts also raises concerns about detecting and preventing the misuse of LLMs~\citep{pagnoni2022threat, mirsky2023threat}. 

One particular problem is the prevalence of AI-generated reviews, and while tools and datasets have been proposed, none have solved the problem completely~\citep{wu2023survey}, which in turn, leads to the erosion of trust in online opinions. 
Furthermore, most of the research so far has focused primarily on English, with very little work dedicated to other languages. 
Our study aims to fill in these gaps and provide novel insights into multilingual LLM-generated hotel reviews. 
We analyze different types of human-interpretable features, such as linguistic style, writing structure, topics, and psycholinguistic markers, along with baselines across multiple models. 
We hope our research will help organizations leverage NLP to combat the use of LLMs in scenarios where genuine, human-generated content is highly valued, such as customer reviews on platforms like Booking, Yelp, and Amazon.

Specifically, our paper aims to answer two main research questions.
\begin{description}[]
\item [RQ1:] {\bf What are the linguistic markers (syntactic and lexical) of multilingual LLM-generated reviews when compared to human-authored reviews?} %
\vspace{-0.3em}
\item [RQ2:] {\bf Which factors influence multilingual deception detection performance?}
\end{description}

The paper makes the following contributions.
First, {\bf we compile and share a multilingual dataset of 10,000 real and 10,000 AI-generated fake hotel reviews, balanced across ten languages:} Chinese, English, French, German, Italian, Korean, Romanian, Russian, Spanish, and Turkish, as well as across ten locations and different sentiment polarities. To the best of our knowledge, this is the first dataset of multilingual reviews at this scale.
Second, using this dataset, {\bf we conduct extensive linguistic style and lexical analyses to compare the AI-generated deceptive hotel reviews with the real human-written hotel reviews.} 
Finally, {\bf we explore the effectiveness of different models for multilingual deception detection} in hotel reviews across three dimensions: sentiment, location, and language.

\section{Related Work}

\paragraph{LLM-generated Text Detection.}
The powerful generation capabilities of LLMs has made it challenging for humans to differentiate between LLM-generated and human-written texts~\cite{jakesch2023human}.
This led to extensive research being conducted on developing models for detecting LLM-generated text, including fine-tuned LMs~\citep{jawahar2020automatic, guo2023close}, zero-shot methods~\citep{solaiman2019release, ippolito2019automatic}, watermarking techniques~\citep{kirchenbauer2023watermark, kirchenbauer2023reliability}, adversarial learning methods~\citep{hu2023radar, koike2024outfox}, LLMs as detectors~\citep{bhattacharjee2024fighting, yu2023gpt}, and human-assisted methods~\citep{Dou2021IsGT}. 

Fine-tuning LMs, in particular Roberta~\cite{Liu2019RoBERTaAR}, has had great success in binary classification settings~\citep{fagni2021tweepfake, radford2019language}, and therefore, we also adopt it in our work.
On average, these models yielded a 95\% accuracy, outperforming zero-shot, and watermarking methods and showing resilience to various attacks within in-domain settings. However, just like other encoder-based fine-tuning approaches, these models lack robustness when dealing with cross-domain or unforeseen data~\cite{bakhtin2019real, antoun2023towards}.

Our work falls under the category of black-box modeling, as described in a recent survey of AI language detectors by \citet{tang2023science}. We are interested in the outputs of LLMs, rather than the specific details of a model's contents or design. This approach allows us to focus on the differences between human and AI-generated texts, instead of the particular implementation details of models. \vspace{-0.5em}

\paragraph{Human vs. LLM-generated Text.}
There have been several efforts to study the differences between AI and human-generated text~\citep{Sadasivan2023CanAT, jakesch2023human, markowitz2024linguistic, markowitz2024generative} 

In the context of deception, \citet{Giorgi2023ISL} and \citet{markowitz2024generative} argue that AI-generated text is \textit{inherently deceptive} when describing human experiences like writing reviews because the system is not grounded in material world experiences. At the same time, \citet{Giorgi2023ISL} and \citet{markowitz2024generative} are the closest to our work. They use ChatGPT to generate hotel reviews and compare them to human-written deceptive and truthful hotel reviews from TripAdvisor. This data is collected by \citet{ott2011finding} from 20 hotels in Chicago, IL, USA.
\citet{markowitz2024generative} find that AI-generated text has a more analytic style and is more affective, more descriptive, and less readable than human-written text, while 
\citet{Giorgi2023ISL} find that human-written text is more diverse in its expressions of personality than AI-generated text.
We replicate their style analysis findings and extend the data collection and analysis to nine other languages and ten global hotel locations. 
Unlike prior work, we do not analyze deceptive human reviews. In addition, we also analyze the role of sentiment, language, and location in deception performance and implicitly in the quality of the AI-generated reviews. \vspace{-0.5em}

\paragraph{Multilingual LLM-generated Text Detection.}
While several multilingual models~\cite{tulchinskii2024intrinsic, mitchell2023detectgpt, antoun2023towards, guo2023close} and datasets~\citep{wang2023m4, wiki-40b} have been proposed, \citet{Wu2023ASO}, in their comprehensive survey on LLM-generated text, address the need for multilingual datasets and models to facilitate the evaluation of text detectors generated by LLMs across different languages. Addressing them is essential for the usability and fairness of detectors for LLM-generated text. 
In our work, we address this gap, by providing a dataset, analysis, and classification models for 10 languages.\vspace{-0.2em}

\section{The \textsc{MAiDE-up} Dataset}
To answer our research questions, we compile a novel dataset, which we refer to as \textsc{MAiDE-up} - Multilingual Ai-generateD fakE reviews. 
\textsc{MAiDE-up} contains a total of 20,000 hotel reviews: 10,000 are real, human-written, and 10,000 are fake, LLM-generated. The reviews are balanced across language, location, and sentiment.
We outline our process for collecting real and fake reviews below.
\vspace{-0.5em}
\subsection{Real Hotel Reviews}
We collect 10,000 hotel reviews from Booking.com, \footnote{\url{booking.com}} which is one of the largest marketplaces for online travel bookings.
The data is balanced per language, location, and sentiment. Finally, we ensure data quality through automatic and manual assessments.
\vspace{-1.6em}
\paragraph{Languages.} 
The dataset is balanced across ten languages: \textit{Chinese, English, French, German, Italian, Korean, Romanian, Russian, Spanish}, and \textit{Turkish}. 
We automatically web-crawl Booking.com for each of the ten languages. Additional details on how we automate this process can be found in \Cref{real}.
\vspace{-0.6em}
\paragraph{Locations.} We collect reviews from hotels located in popular capital cities: \textit{Ankara, Beijing, Berlin, Bucharest, Madrid, New Delhi, Paris, Rome, Seoul}, and \textit{Washington}. 
Most of the cities are selected to be the capitals of countries where the official language is one of the 10 languages (New Delhi and Moscow are the exceptions).
To ensure the collection of an equal number of reviews for each language, particularly for \textit{Chinese, Korean, Romanian}, and \textit{Turkish}, where the number of reviews per hotel is often limited, we identify 250 hotels in each city and collect up to 50 reviews per hotel. 

\begin{figure}[h!]
    \centering
    \includegraphics[width=0.8\linewidth]{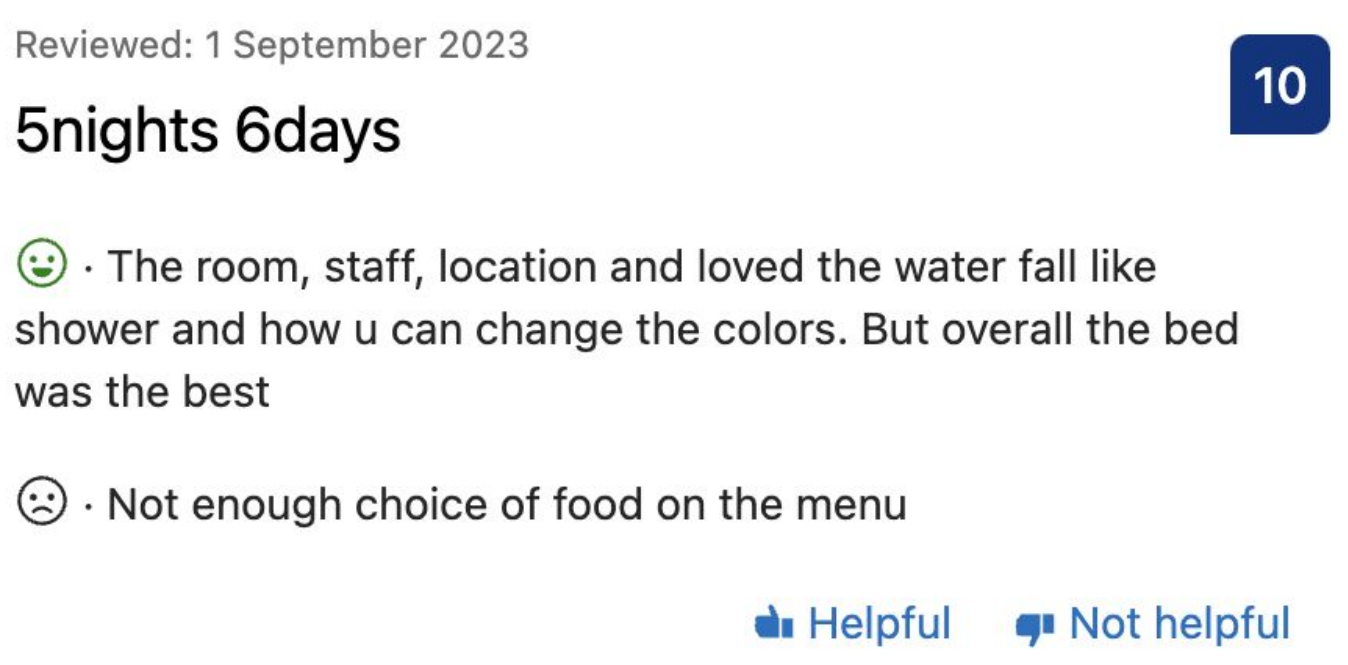}
    \caption{An example of an English positive review, rated with a score of 10, that contains both ``upside'' and ``downside'' sections. The reviewer can choose to write in just one or both sections.\vspace{-2em}}
    \label{fig:sample-review}
\end{figure}

\vspace{-0.3em}
\paragraph{Sentiment.}
The Booking.com platform provides \textit{review scores} from 1 to 10, where a score of 1-6 represents a negative review and 7-10 is a positive review.\footnote{Even though it is marked as ‘Pleasant’ on Booking.com, a score of 6 is considered negative on the platform.} We collect a balanced number of positive and negative reviews for each language, i.e., 500 positive and 500 negative for each language.
The platform provides a specific review format consisting of two parts, ``upside'' and ``downside'', to allow users to separate their positive and negative feedback. The ``upside'' and ``downside'' sections are optional, as the reviewer can write in just one or both sections. 
\Cref{fig:sample-review}, shows a review example of a positive review, rated with a score of 10, that contains both ``upside'' and ``downside'' sections.

\vspace{-0.6em}
\paragraph{Quality Assurance.}
We automatically verify each review's language and filter out reviews in a different language than the one crawled for (more information in \Cref{real}).
To further ensure the data quality, ten native speakers manually verified 50 random reviews each. Specifically, they verified the review's syntax and semantics and ensured the sentiment aligned with the content for the ``upside'' and ``downside'' parts of the review, as well as for the review overall.
One potential concern is the possibility of fake human-written reviews. Booking.com addresses this by combining specialized personnel and automated systems to detect and remove fake reviews.
\vspace{-0.6em}

\subsection{LLM-generated Hotel Reviews} 

We generate 10,000 hotel reviews with GPT-4.\footnote{\url{https://platform.openai.com/docs/guides/text-generation/chat-completions-api}} The generated reviews follow the same distribution across language, location, and sentiment, as the real hotel reviews, as described in section 3.1.
We use GPT-4 because it is one of the largest LLMs available and has been demonstrated to effectively emulate human texts~\citep{achiam2023gpt}.

\subsubsection{Prompt Design and Robustness}
GPT-4 takes a prompt as input, which is comprised of a list of \textit{message} objects, and returns one generated hotel review as output. 
We use \textit{messages}, which are more interactive and dynamic compared to the classical prompt-style. Specifically, we use messages with two properties: \textit{role} and \textit{content}: the role takes one of three values: ``system'', ``user'' or ``assistant'', while the content contains the text of the message.\footnote{\url{https://help.openai.com/en/articles/7042661-chatgpt-api-transition-guide}}

The prompt is first formatted with a ``system message'' role, which sets the behavior of the model. This is followed by two rounds of conversations between the roles of ``assistant'' and ``user'' in a \textit{few-shot prompting} technique.
Finally, we use a ``user message'' to prompt the model to generate a hotel review in a specified language, location (hotel name and capital city), and sentiment. 
\vspace{-0.3em}
\paragraph{System Prompt.}
We find that we can obtain high-quality responses with additional context in our prompts.
Therefore, we instruct the model to be a well-traveled native \{{language}\} tourist in the "system message". The \{{language}\} placeholder is replaced by the language name of the hotel reviews we aim to collect. The instructions also contain information about the language, location (hotel name and city) and the sentiment of the review, as well as the output format, as illustrated below: 

\begin{quote}
\vspace{-0.8em}
{
\small
\textit{You are a well-traveled native \{language\} tourist, working on writing hotel reviews of hotels you have stayed in. Given hotel name, city name, language, and sentiment, you write a hotel review comprised of upside and downside. Then you give an overall review score, an integer ranging from 1 to 10 where the score larger than 6 indicates positive experience, otherwise negative experience. 
You always output a JSON containing the following keys: ``Upside\_Review'', ``Downside\_Review'', ``Review\_Score''. 
Reviews are always in consistent styles, tone, sentence structure.}
}
\end{quote}
\vspace{-1em}
\paragraph{User Prompt.}
To increase the diversity and robustness of the generated data, we collect 36 multilingual ``user messages''. Specifically, we ask ten native speakers to each write four ``user messages'': two in their own language and two in the corresponding \textit{English} translation. The guidelines given to the annotators are shown in \Cref{llm}.

We use the ``user messages'' to prompt GPT-4 to generate a hotel review with a specified language, hotel location, and sentiment. Below is an example of a ``user message'' in \textit{Spanish} and its \textit{English} translation. We show all the multilingual ``user messages'' in the Appendix \Cref{tab:all-prompts}.

\begin{quote}
{\small
\vspace{-0.4em}
\textit{¿Podés escribir un comentario positivo en {L} sobre el hotel {H} de {C}?}
\vspace{-0.5em}
\textit{Can you please write a positive review in {L} for the hotel {H} located in {C}?}
\vspace{-0.4em}}
\end{quote}

\paragraph{Few-shot Prompting.}
We use the conversations between ``user" and ``assistant'' when generating data. The ``user messages'' are randomly selected from our multilingual ``user messages'' and the ``assistant messages" are randomly extracted from our collected real hotel reviews. 
The answer to the last ``user message'' is automatically generated by the ``assistant message'' and used to collect the GPT-4 generated hotel reviews. 
\vspace{-0.4em}
\paragraph{Quality Assurance.}
To ensure the quality of our generated data, we conduct sanity checks by asking native speakers to review approximately 100 hotel reviews in their respective languages. The reviews are checked for readability, syntax errors, and style. Based on the feedback, we find that some of the generated \textit{Chinese} reviews contain nonsensical phrases. Additionally, reviews in other languages such as \textit{Romanian, Korean}, and \textit{Spanish} have a formal tone.
\Cref{tab:review-examples} displays random reviews in \textit{English}, both real and generated, with positive and negative sentiments.

\paragraph{Cost.}
We generated 10,000 posts in 10 languages for a total cost of 250-300 dollars
(0.03 per 1K input tokens and 0.06 per 1K output tokens).

\begin{table}[h]
\centering
\small
{\renewcommand{\arraystretch}{1.1}%
    \begin{tabular}{p{7cm}}
    \toprule
\textbf{Real Hotel Reviews} \\
    \midrule
    $\color{green}\textbf{+}$ \textit{Nicely furnished room and nicely decorated lobby. Room service is affordable and the receptionists, especially Mr. Omar, are usually eager to help with any queries; N/A} \\
    $\color{red}\textbf{-}$ \textit{The location was very central and the staff was nice and helped us.; The room smelled like cigarettes and there was mold in the bathroom}\\
    \midrule
  \textbf{LLM-generated Hotel Reviews} \\
  \midrule
    $\color{green}\textbf{+}$ \textit{The staff was extremely helpful and accommodating. Clean and well-furnished rooms. Central location with easy access to public transport.; The breakfast offerings could be more varied.} \\ 
    $\color{red}\textbf{-}$ \textit{Nothing. Horrible experience.; Bad customer service, rooms were not clean and the food was below average.} \\ 
    \bottomrule
    \end{tabular}
    }
    \caption{Random review samples with positive ($\color{green}\textbf{+}$) and negative ($\color{red}\textbf{-}$) sentiments in English. ``Upside'' and ``Downside'' parts are separated by ``;''. Multilingual sampled reviews are shown in Appendix Tables \ref{tab:review-examples-all1} and \ref{tab:review-examples-all2}.\vspace{-1em}}
    \label{tab:review-examples}
\end{table}

\section{Multilingual Analyses of Real and LLM-generated Hotel Reviews}
Using our dataset, we conduct extensive linguistic analysis to compare the AI-generated fake hotel reviews with the real human-written hotel reviews.\vspace{-0.5em}

\paragraph{Analytic Writing.} This is an index that measures the complexity and sophistication of the writing, which can be an indicator of advanced thinking. This technique has been used in various fields, including persuasion~\citep{markowitz2020putting}, analysis of political speeches~\citep{jordan2019examining}, and gender studies~\citep{meier2020stereotyping}, among others.
The formula for analytic writing is $[articles + prepositions - pronouns - auxiliary verbs - adverb - conjunctions - negations]$ from LIWC scores~\citep{jordan2019examining, pennebaker2014small}.
We use Linguistic Inquiry and Word Count (LIWC)~\citep{Pennebaker2007LinguisticIA,Pennebaker2015TheDA}, a gold-standard text analysis tool, to obtain the categories in the index formula for the following languages: \textit{Chinese}, \textit{English}, \textit{French}, and \textit{Spanish}.\footnote{While versions for Korean and Turkish are listed as available upon request, we were unable to obtain them.} 
Since they are translations for different versions of the LIWC lexicon, e.g., 2001, 2007, and 2015, we first align them using the intersection with the 2015 English version. 
We find that \textbf{AI-generated texts in \textit{English} are more complex than real texts}, which aligns with our observations from data quality checks. The results for \textit{Chinese}, \textit{French}, and \textit{Spanish} are not statistically significant (see \Cref{tab:style}).\vspace{-0.5em}

\paragraph{Descriptiveness.} The descriptiveness of a text can be measured by its ratio of adjectives, as texts with high rates of adjectives tend to be more elaborate and narrative-like compared to texts with low rates of adjectives.~\citep{chung2008revealing} Additionally, adjectives are often used in false speech, making them a key marker of deceitful language~\citep{johnson1981reality}.
We measure the ratio of adjectives using the multilingual library \texttt{textdescriptives} from \cite{hansen2023textdescriptives}.\footnote{\url{https://github.com/HLasse/TextDescriptives}}
For \textit{Turkish} we use \texttt{HuggingFace} from \citet{altinok-2023-diverse}.\footnote{\url{https://huggingface.co/turkish-nlp-suite}}
In \Cref{tab:style}, we show that \textbf{AI-generated text is usually more descriptive than real text}. The only exceptions are \textit{German} reviews, where the real text is more descriptive, and \textit{Korean} reviews, where the difference is not significant.\vspace{-0.5em}  

\paragraph{Readability.} The readability of a text is reflected not only by its word count,  but also by the word complexity, with e.g., longer words being more difficult to read and understand.  We use the Flesch Reading Ease metric~\citep{flesch1948new}, which counts the number of words per sentence and syllables per word. This metric is used to assess the structural complexity of language patterns in various texts, such as scientific articles~\citep{markowitz2016linguistic}, online petitions~\citep{markowitz2023instrumental}, and social media data~\citep{hubner2022scientist}. 
We use the multilingual library \texttt{textdescriptives} to compute the Flesch Reading Ease metric and the word count. 
In \Cref{tab:style}, we find that \textbf{AI-generated text is usually less readable and more wordy than real text}. The only exceptions are \textit{German} and \textit{Russian}, with no significant differences.  \vspace{-0.5em}

\begin{table*}[h!]
\resizebox{\textwidth}{!}{%
\begin{tabular}{c| ccc |ccc| ccc| ccc}
\toprule
& \multicolumn{3}{c|}{\textbf{Analytic writing}} 
& \multicolumn{3}{c|}{\textbf{Descriptiveness}}
& \multicolumn{3}{c|}{\textbf{Readability}}
& \multicolumn{3}{c}{\textbf{Word Count}}
\\
\midrule
\multirow{1}{*}{\textbf{Lang}} &
  \multicolumn{1}{c}{\textbf{Real}} &
  \multicolumn{1}{c}{\textbf{Gen}} &
  \multicolumn{1}{c|}{\textbf{Diff}} &
  \multicolumn{1}{c}{\textbf{Real}} &
  \multicolumn{1}{c}{\textbf{Gen}} &
   \multicolumn{1}{c|}{\textbf{Diff}} &
  \multicolumn{1}{c}{\textbf{Real}} &
  \multicolumn{1}{c}{\textbf{Gen}} &
   \multicolumn{1}{c|}{\textbf{Diff}} &
  \multicolumn{1}{c}{\textbf{Real}} &
  \multicolumn{1}{c}{\textbf{Gen}} &
  \multicolumn{1}{c}{\textbf{Diff}}\\
\midrule
\worldflag[length=0.5cm, width=0.3cm]{CN} & 6.2 $\pm$ 7.0 & 6.5 $\pm$ 3.6 & \cellcolor{white}0.3 & 2.6 $\pm$ 3.9$^*$ & 4.2 $\pm$ 3.5$^*$ & \cellcolor{teal1}1.6 &  - & - &- &  62.1 $\pm$ 92.9$^*$ & 79.4 $\pm$ 41.2$^*$  & \cellcolor{teal1}17.3\\
\midrule
\worldflag[length=0.5cm, width=0.3cm]{US} & 11.9 $\pm$ 6.8$^*$ & 18.6 $\pm$ 5.2$^*$ & \cellcolor{teal1}6.7 &  13.7 $\pm$ 8.6$^*$ & 15.5 $\pm$ 5.5$^*$ & \cellcolor{teal1}1.8 & 66.3 $\pm$ 29.0$^*$ & 54.6 $\pm$ 17.1$^*$ & \cellcolor{olive1}-11.7 & 49.2 $\pm$ 54.7$^*$ & 55.7 $\pm$ 28.5$^*$& \cellcolor{teal1}6.5\\ 
\midrule
\worldflag[length=0.5cm, width=0.3cm]{FR} & 21.0 $\pm$ 10.1 & 20.8 $\pm$ 6.2& \cellcolor{white}-0.2& 13.9 $\pm$ 9.3$^*$ & 16.2 $\pm$ 6.7$^*$& \cellcolor{teal1}2.3 & 51.3 $\pm$ 29.9$^*$ & 43.9 $\pm$ 15.8$^*$& \cellcolor{olive1}-7.4 & 39.6 $\pm$ 43.5$^*$ & 47.5 $\pm$24.2$^*$& \cellcolor{teal1}7.9\\ 
\midrule
\worldflag[length=0.5cm, width=0.3cm]{DE} & - & - & \cellcolor{white}-& 7.0 $\pm$ 7.4$^*$ & 5.4 $\pm$ 5.5$^*$& \cellcolor{olive1}-1.6 & 28.2 $\pm$ 32.3 & 27.2 $\pm$ 19.3& \cellcolor{white}-1 & 41.2 $\pm$ 43.0$^*$ & 47.0 $\pm$ 22.4$^*$& \cellcolor{teal1}5.8\\
\midrule
\worldflag[length=0.5cm, width=0.3cm]{IT} &  -& -& \cellcolor{white} - & 15.4 $\pm$ 10.2$^*$ & 19.3 $\pm$ 7.6$^*$ & \cellcolor{teal1}3.9 & -9.4 $\pm$ 32.0$^*$ & -13.7 $\pm$ 17.9$^*$ & \cellcolor{olive1}-4.3 & 37.9 $\pm$ 41.7$^*$ & 45.0 $\pm$ 23.2$^*$& \cellcolor{teal1}7.1\\
\midrule
\worldflag[length=0.5cm, width=0.3cm]{RO} &  -&-& \cellcolor{white} -& 11.3 $\pm$ 9.1$^*$ & 15.5 $\pm$ 6.3$^*$& \cellcolor{teal1} 4.2 & 8.4 $\pm$ 36.7$^*$ & -0.6 $\pm$ 19.3$^*$ & \cellcolor{olive1} -9& 39.6 $\pm$ 46.3$^*$ & 44.1 $\pm$ 22.1$^*$& \cellcolor{teal1} 4.5\\
\midrule
\worldflag[length=0.5cm, width=0.3cm]{KR} & - & -& \cellcolor{white} -& 6.0 $\pm$ 6.4 & 5.7 $\pm$ 4.3 & \cellcolor{white} -0.3& - & - & \cellcolor{white} -& 30.1 $\pm$ 31.7$^*$ & 32.6 $\pm$ 14.1$^*$& \cellcolor{teal1} 2.5\\
\midrule
\worldflag[length=0.5cm, width=0.3cm]{RU} &  -& -& \cellcolor{white} -& 13.7 $\pm$ 8.7$^*$ & 18.9 $\pm$ 6.0$^*$ & \cellcolor{teal1} 5.2& -8.2 $\pm$ 35.1$^*$ & -16.8 $\pm$ 20.0$^*$ & \cellcolor{olive1} -8.6& 42.8 $\pm$ 45.5 & 40.7 $\pm$ 20.5& \cellcolor{white} -1.1\\
\midrule
\worldflag[length=0.5cm, width=0.3cm]{ES} & 12.9 $\pm$ 7.0 & 15.9 $\pm$ 4.1& \cellcolor{white} 3& 11.5 $\pm$ 8.6$^*$ & 14.9 $\pm$ 5.4$^*$ & \cellcolor{teal1} 3.4& 19.9 $\pm$ 27.6$^*$ & 11.0 $\pm$ 17.1$^*$ & \cellcolor{olive1}-8.9& 39.0 $\pm$ 40.5$^*$ & 48.0 $\pm$ 23.1$^*$& \cellcolor{teal1} 9\\
\midrule
\worldflag[length=0.5cm, width=0.3cm]{TR} & - & -& \cellcolor{white} -& 13.6 $\pm$ 9.5$^*$ & 14.4 $\pm$ 6.0$^*$ & \cellcolor{teal1} 0.8& - & - & \cellcolor{white} -& 26.4 $\pm$ 25.3$^*$ & 32.8 $\pm$ 14.7$^*$& \cellcolor{teal1} 6.4\\
\midrule
Average & 13$\pm$ 7.7& 15.4$\pm$4.7& \cellcolor{teal1}2.4 & 10.8 $\pm$ 8.1 & 13 $\pm$ 5.6 & \cellcolor{teal1}2.1 & 22.3 $\pm$ 31.8& 15 $\pm$ 18.1& \cellcolor{olive1}-7.2& 40.7 $\pm$ 46& 47.2 $\pm$ 23& \cellcolor{teal1}6.5\\
\bottomrule
\end{tabular}%
}
\caption{To what degree is AI-generated text different from real text in terms of analytic writing, descriptiveness, readability, and word count? We compute the mean and standard deviation for all the reviews, across each language.
We mark ($^*$) when the difference between real and generated data is statistically significant, based on the Student t-test~\citep{student1908probable} with p-value < 0.05. The significant differences are indicated in \textit{{\color{teal1}teal}} when the generated data scores are higher than real data scores, and in \textit{{\color{olive1}olive}} otherwise.\vspace{-1em}}
\label{tab:style}
\end{table*}

\paragraph{Topic Modeling.}
We compute the most prevalent topics and their keywords with a multilingual pipeline from Scattertext~\citep{kessler-2017-scattertext}. Each review is processed to obtain the most important words with high TF-IDF scores~\cite{ramos2003using}. 
We pre-process the text with \texttt{spaCy} multilingual pipelines to tokenize hotel reviews and lowercase tokens, remove stop words, and lemmatize tokens.\footnote{\url{https://spacy.io/models}}

Comparing real to generated reviews, we find that in multilingual data, the word ``Booking'' is most frequent in real hotel reviews, along with words like ``reception (ro: receptie)'', ``checking'', ``bathroom (ro: baie, es: bagno)'', ``shower''. In contrast, generated hotel reviews contain more words about ``service'', ``comfort (de: komfortabel, ch: \begin{CJK}{UTF8}{mj}舒适\end{CJK})'' and ``room (ro: camerele)''. In English hotel reviews, AI-generated reviews contain more mentions of city names: ``Bucharest'', ``Washington'', ``Ankara'', while real hotel reviews contain more words about ``cleaning'' and ``time''. The topic distribution across real and AI-generated hotel reviews in all languages is shown in \Cref{fig:topic_multi} and \Cref{fig:topic_english} in the Appendix.

\section{Multilingual Deception Detection}
We explore the effectiveness of different models for multilingual deception detection. 
As interpretable baselines, we train and test a Naive Bayes~\citep{Lewis1998NaiveA} and a Random Forest~\citep{ho1995random} classifier.
As our main model, we fine-tune an XLM-RoBERTa~\citep{roberta} classifier, which lacks interpretability but is highly performant. \vspace{-0.5em}

\paragraph{Model Training Setup.}
We use one model for all the multilingual data and also experiment with different training and test data splits. Specifically, we use a \textit{default}
and a \textit{few-shot} train-test split as shown in \Cref{tab:train-test-split}.
Note that the \textit{few-shot} setting closely resembles the real-life scenario when a user who wants to generate fake multilingual reviews has access to little labeled data.

\subsection{Interpretable Baselines}
We extract multiple interpretable features for each review: token counts, TF-IDF scores~\cite{ramos2003using}, as well as the Analytic Writing, Descriptiveness, and Readability scores described in Sec 4. \vspace{-1em}

\paragraph{Random.} The reviews are split equally between real and generated; therefore, a random baseline has an accuracy of 50\% and an F1 score of 0\%.\vspace{-0.5em}

\paragraph{Naive Bayes.} As a simple and interpretable baseline, we train a Naive Bayes classifier~\citep{Lewis1998NaiveA} to distinguish between real and AI-generated reviews. 
We use the Gaussian Naive Bayes model from the \texttt{sklearn}~\citep{scikit-learn} library with the default settings.\vspace{-0.5em}

\paragraph{Random Forest.} We train a Random Forest classifier~\citep{ho1995random} to distinguish between real and AI-generated reviews.
We use the default model from the \texttt{sklearn}~\citep{scikit-learn} library. \vspace{-1em}

\subsection{Main Model}
\paragraph{XLM-RoBERTa.}
We fine-tune XLM-RoBERTa base~\cite{roberta} model from \texttt{HuggingFace}\footnote{\url{https://huggingface.co/FacebookAI/xlm-roberta-base}}.
The model is a multilingual version of RoBERTa~\citep{Liu2019RoBERTaAR}, and is pre-trained on 2.5TB of filtered CommonCrawl~\citep{wenzek-etal-2020-ccnet} data containing 100 languages.

We use a learning rate of $2e-5$ and a batch size of $8$. We train for $5$ epochs and take the best epoch based on validation accuracy. The validation data represents 10\% of the train data set.

\begin{table}
\centering
\small
\setlength{\tabcolsep}{0.3em} %
{\renewcommand{\arraystretch}{1}%
\begin{tabular}{@{\extracolsep{7pt}} l c r c c@{}}
  \toprule 
      &
        \multicolumn{2}{c}{\textbf{\textit{default}}} &
      \multicolumn{2}{c}{\textbf{\textit{few-shot}}} \\
      \cline{2-3} \cline{4-5}
  \multicolumn{1}{l}{} & real &  \multicolumn{1}{c}{gen} &  
  \multicolumn{1}{c}{real} &  \multicolumn{1}{c}{gen}\\
  \midrule
   Train & 8,000 & 8,000 & 100 & 100 \\ 
   Test & 2,000 & 2,000 & 9,900 & 9,900 \\
\bottomrule
      \end{tabular}
      }
      \newline
    \caption{Number of reviews for the experimental data split. 
    The \textit{default} split corresponds to an 80-20\% train-test data split, while the \textit{few-shot} split corresponds to a 1-99\% train-test data split.\vspace{-1em}}
    \label{tab:train-test-split}
\end{table}

\begin{table}[h]
    \centering
    \small
    \setlength{\tabcolsep}{0.7em} %
    {\renewcommand{\arraystretch}{1}%
    \scalebox{0.95}{
    \begin{tabular}{l c c c}
    \toprule
    \textbf{Features} & \textbf{Setup} & \textbf{Accuracy} & \textbf{F1} \\
    \midrule
    \midrule
    \multicolumn{4}{c}{Interpretable Baselines: \sc Naive Bayes / Random Forest} \\
    \midrule
    \midrule
    \multirow{2}{*}{Analytic scores} & {\textit{default}} &  53.3 / 53.9 & 63.2 / 64.3 \\
     & {\textit{few-shot}} & 52.1 / 51.1 & 63.6 / 32.1 \\
    \midrule
    \multirow{2}{*}{Descriptive scores} & {\textit{default}}  & 58.3 / 61.0 & 63.1 / 58.6 \\
    & {\textit{few-shot}} & 58.4 / 57.4 & 57.7 / 57.5\\
     \midrule
     \multirow{2}{*}{Readability scores} & {\textit{default}} & 53.5 / 57.3 & 61.6 / 58.7 \\
     & {\textit{few-shot}} & 53.6 / 53.0 & 61.8 / 57.7\\
    \midrule
     \multirow{2}{*}{Token counts}  & {\textit{default}} & 79.6 / 87.4 & 82.2 / 88.0 \\
     & {\textit{few-shot}} & 54.4 / 62.9 & 54.9 / 66.6\\
     \midrule
     \multirow{2}{*}{TF-IDF scores} & {\textit{default}} &  84.5 / 87.7 & 83.9 / 86.9 \\
    & {\textit{few-shot}} & 64.6 / 67.1 & 69.6 / 57.1\\
     \midrule
    \multirow{2}{*}{All scores} & {\textit{default}} &  84.3 / \textbf{89.3}  & 84.1 / \textbf{89.2}\\
    & {\textit{few-shot}}  & 66.3 / 70.6 & 70.5 / 66.8\\
    \midrule
    \midrule
    \multicolumn{4}{c}{Main Model: \sc XLM-RoBERTa} \\
    \midrule
    \midrule
      - &{\textit{default}} & \textbf{94.8} & \textbf{94.9}\\
     - & {\textit{few-shot}} & 76.6 & 80.1\\
    \midrule
    \midrule
    Human & - & 71.5 & 69.1\\
    \bottomrule
    \end{tabular}
    }
    }
    \caption{Classification test results with the \textit{few-shot} and \textit{default} setups over all languages. \vspace{-1em}}
    \label{tab:eval_all}
\end{table}

\begin{figure}[h!]
    \centering
\includegraphics[width=0.7\linewidth]{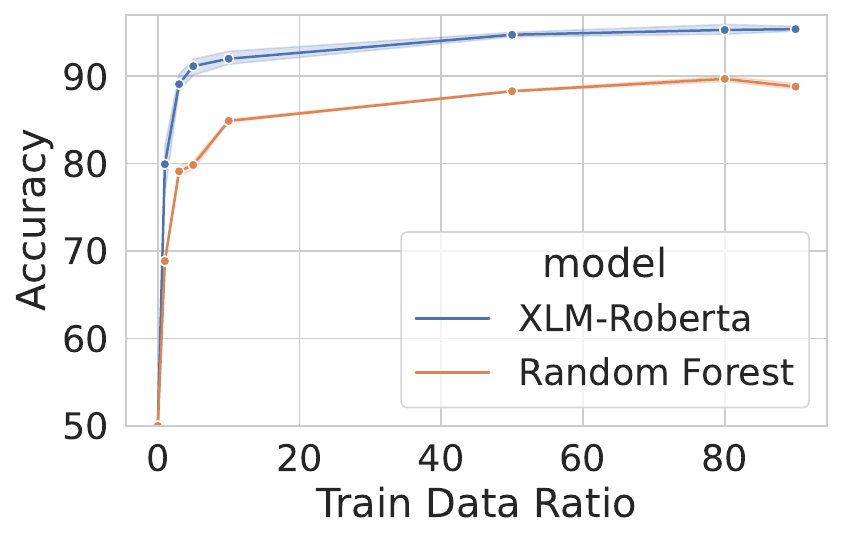}
    \caption{Accuracy measured with XLM-RoBERTa and best Random Forest model on different ratios of training data. The accuracy plateaus at $~$10\%, i.e., 2,000 reviews.\vspace{-1em}} 
    \label{fig:eval_ratio}
\end{figure}

\begin{table*}[h!]
\centering

\resizebox{\textwidth}{!}{%
\begin{tabular}{p{13cm}|p{13cm}}
\toprule
\textbf{Most salient words for the Real reviews} & \textbf{Most salient words for the Generated reviews} \\ \midrule
\textbf{NUM:} 20, 10, 15, 100, 30, 50, tres (three, \worldflag[length=0.5cm, width=0.3cm]{ES}), dos (two, \worldflag[length=0.5cm, width=0.3cm]{ES}) &  
\textbf{VB \& ADV:} \fontencoding{T2A}\selectfont{желать}  (want, \worldflag[length=0.5cm, width=0.3cm]{RU}), \begin{CJK}{UTF8}{mj}또한\end{CJK} (also, \worldflag[length=0.5cm, width=0.3cm]{KR}), lasciava (left, \worldflag[length=0.5cm, width=0.3cm]{IT}), \begin{CJK}{UTF8}{mj}않았습니다\end{CJK} (however, \worldflag[length=0.5cm, width=0.3cm]{KR}), ayudar (help, \worldflag[length=0.5cm, width=0.3cm]{ES})
\\
\midrule
\textbf{ADJ:}
yüzlü (faced, \worldflag[length=0.5cm, width=0.3cm]{TR}),
ok (-, \worldflag[length=0.5cm, width=0.3cm]{US}),   
cok (a lot, \worldflag[length=0.5cm, width=0.3cm]{TR}), 
min (-,  \worldflag[length=0.5cm, width=0.3cm]{US}),  
noi (new,  \worldflag[length=0.5cm, width=0.3cm]{RO}),  
\begin{CJK}{UTF8}{mj}좋아요\end{CJK} (great, \worldflag[length=0.5cm, width=0.3cm]{KR}) & 
\textbf{ADJ:} \fontencoding{T2A}\selectfont{отеля}{лучшего} (the best, \worldflag[length=0.5cm, width=0.3cm]{RU}), enttäuschend (disappointing, \worldflag[length=0.5cm, width=0.3cm]{DE}), \begin{CJK}{UTF8}{mj}만족스럽지\end{CJK} (not satisfied,  \worldflag[length=0.5cm, width=0.3cm]{KR}), limpias(clean,  \worldflag[length=0.5cm, width=0.3cm]{ES}), 
\fontencoding{T2A}\selectfont{отеля}{дружелюбным} (friendly,  \worldflag[length=0.5cm, width=0.3cm]{RU}), 
dispuesto (willing,  \worldflag[length=0.5cm, width=0.3cm]{ES}), \begin{CJK}{UTF8}{mj}위치해\end{CJK} (located,  \worldflag[length=0.5cm, width=0.3cm]{KR}),  
prietenos (friendly,  \worldflag[length=0.5cm, width=0.3cm]{RO}),  
\begin{CJK}{UTF8}{mj}편리했습니다\end{CJK} (convenient,  \worldflag[length=0.5cm, width=0.3cm]{KR}), negative (negative,  \worldflag[length=0.5cm, width=0.3cm]{ES}),
lent (slow,  \worldflag[length=0.5cm, width=0.3cm]{RO})\\
\midrule
\textbf{NOUN:} check (-,  \worldflag[length=0.5cm, width=0.3cm]{US}), 
\begin{CJK}{UTF8}{mj}가성\end{CJK} (cost-effectiveness,  \worldflag[length=0.5cm, width=0.3cm]{KR}), 
habitacion (room,  \worldflag[length=0.5cm, width=0.3cm]{ES}),    
ubicacion (location,  \worldflag[length=0.5cm, width=0.3cm]{ES}),  
güler (laughs,  \worldflag[length=0.5cm, width=0.3cm]{TR}),
euro (-,  \worldflag[length=0.5cm, width=0.3cm]{US}), 
booking (-,  \worldflag[length=0.5cm, width=0.3cm]{US}),      
\begin{CJK}{UTF8}{mj}그런지\end{CJK} (grunge,  \worldflag[length=0.5cm, width=0.3cm]{KR}),  
terlik (slipper,  \worldflag[length=0.5cm, width=0.3cm]{TR}),
kahvalti (breakfast,  \worldflag[length=0.5cm, width=0.3cm]{TR}),
minuti (minutes,  \worldflag[length=0.5cm, width=0.3cm]{IT}),
\fontencoding{T2A}\selectfont{минут} (minutes,  \worldflag[length=0.5cm, width=0.3cm]{RU}) 
& \textbf{NOUN:} serviciul (service,  \worldflag[length=0.5cm, width=0.3cm]{RO}), 
seçenekleri (options,  \worldflag[length=0.5cm, width=0.3cm]{TR}),  
bağlantısı (connection,  \worldflag[length=0.5cm, width=0.3cm]{TR}), 
wifiul (the wifi,  \worldflag[length=0.5cm, width=0.3cm]{RO}),
oraș (city,  \worldflag[length=0.5cm, width=0.3cm]{RO}), 
kalitesi (quality, \worldflag[length=0.5cm, width=0.3cm]{TR}),  
\begin{CJK}{UTF8}{mj}호텔의\end{CJK} (hotel,  \worldflag[length=0.5cm, width=0.3cm]{KR}) \\
\bottomrule
\end{tabular}
}
\caption{Top 20 most salient features for real and generated reviews from the best Random Forest model on the \textit{default} train-test split.
Each word is accompanied by its English translation, part of speech, and language.\vspace{-1em}}
\label{tab:feature_importances}
\end{table*}

\section{Evaluation} 
We show the Accuracy and F1 results for all the models in \Cref{tab:eval_all}.
The model with the best detection performance is XLM-RoBERTa, with an accuracy of $94.8\%$ on the \textit{default} train-test data split and $76.6\%$ accuracy on \textit{few-shot} train-test split. 

Among all the interpretable models, Random Forest with all the features achieves the best accuracy of $89.3$ on the \textit{default} train-test data split and $70.6\%$ accuracy on the \textit{few-shot} train-test data split. \textbf{The features that contribute the most to the performance increase are TF-IDF scores, followed by Token counts, Descriptiveness, Readability, and Analytic scores.} Note that missing data for several languages, as seen in \Cref{tab:style}, impacts the performance of Analytic scores and Readability scores. To handle the missing values, we replace them with the mean of the present values. 
Even though they contribute to performance increase along with the TF-IDF scores, we find that the analytic writing, descriptiveness, and readability scores are not sufficient to accurately distinguish between real and LLM-generated hotel reviews.

Both models have a fairly high \textit{few-shot} performance, indicating that they are capable of learning this task from very few examples, i.e., 200 reviews balanced across 10 languages.
Furthermore, \Cref{fig:eval_ratio} shows that \textbf{both models learn considerably from just 2,000 reviews} (train data ratio of 10\%), and increasing the training data does not lead to high performance gains.\vspace{-0.5em}

\begin{figure}[h!]
    \centering
\includegraphics[width=\linewidth]{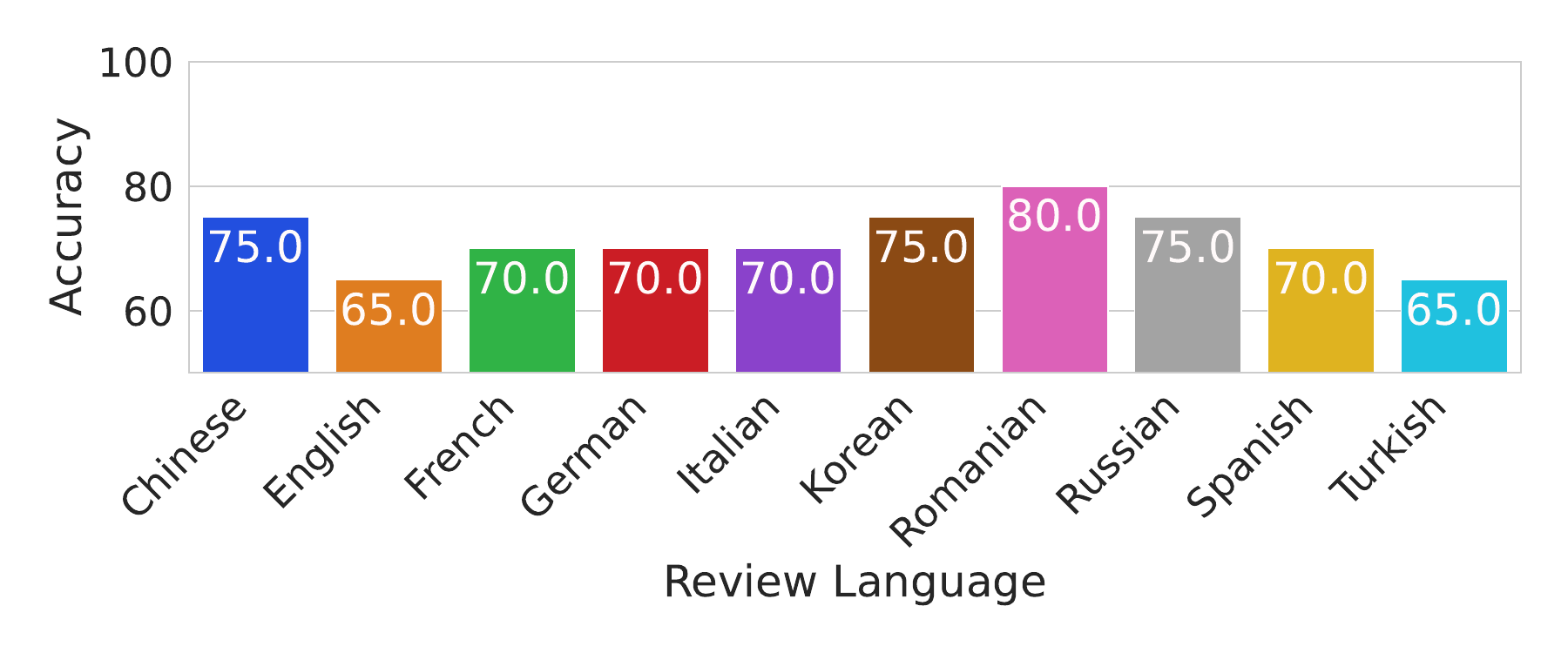}
    \caption{Human accuracy per review language.} 
    \label{fig:eval_human}
\end{figure}

\paragraph{Human Evaluation.} We ask ten native speakers to manually classify 200 random reviews (100 real, 100 generated) across all languages as deceptive or not. The accuracy per language is shown in \Cref{fig:eval_human}.  The $71.5\%$ overall accuracy suggests that \textbf{humans find it moderately difficult to distinguish between real and generated hotel reviews}. 
Indeed, the annotators mention that the task is challenging, and they tend to label more complex and formal reviews as AI-generated.
In \Cref{tab:eval_all}, human performance is comparable to the \textit{few-shot} learning model performance. 
However, \textbf{all the models trained on more data (\textit{default} split) significantly outperform humans}, indicating that they are helpful for deception detection.

\subsection{Interpretability Analysis}
Table \ref{tab:feature_importances} shows the top 20 most salient features of the best Random Forest classifier in the \textit{default} training setup. We observe that \textit{Korean, Spanish,} and \textit{Turkish} words appear most frequently, suggesting they have the strongest impact on deception detection performance.
Additionally, for both real and generated reviews, words related to the hotel topic are most salient: \textit{cost, room, location, booking, breakfast, wifi, service}. Compared to generated reviews, real reviews tend to have more numerals as salient words, while generated reviews have more verbs, adverbs, and adjectives. This finding aligns with our previous discovery that when measuring descriptiveness scores, generated reviews are typically more descriptive than real reviews. Finally, the nouns in real reviews tend to be more diverse than those in generated reviews: \textit{cost-effectiveness, laughs, grunge, slipper}. We encourage further research to delve more deeply into the lexical variances between real and generated multilingual reviews using our dataset.

\subsection{Ablations per Language, Location, and Sentiment}
\begin{figure}[h!]
    \centering
    \includegraphics[width=\linewidth]{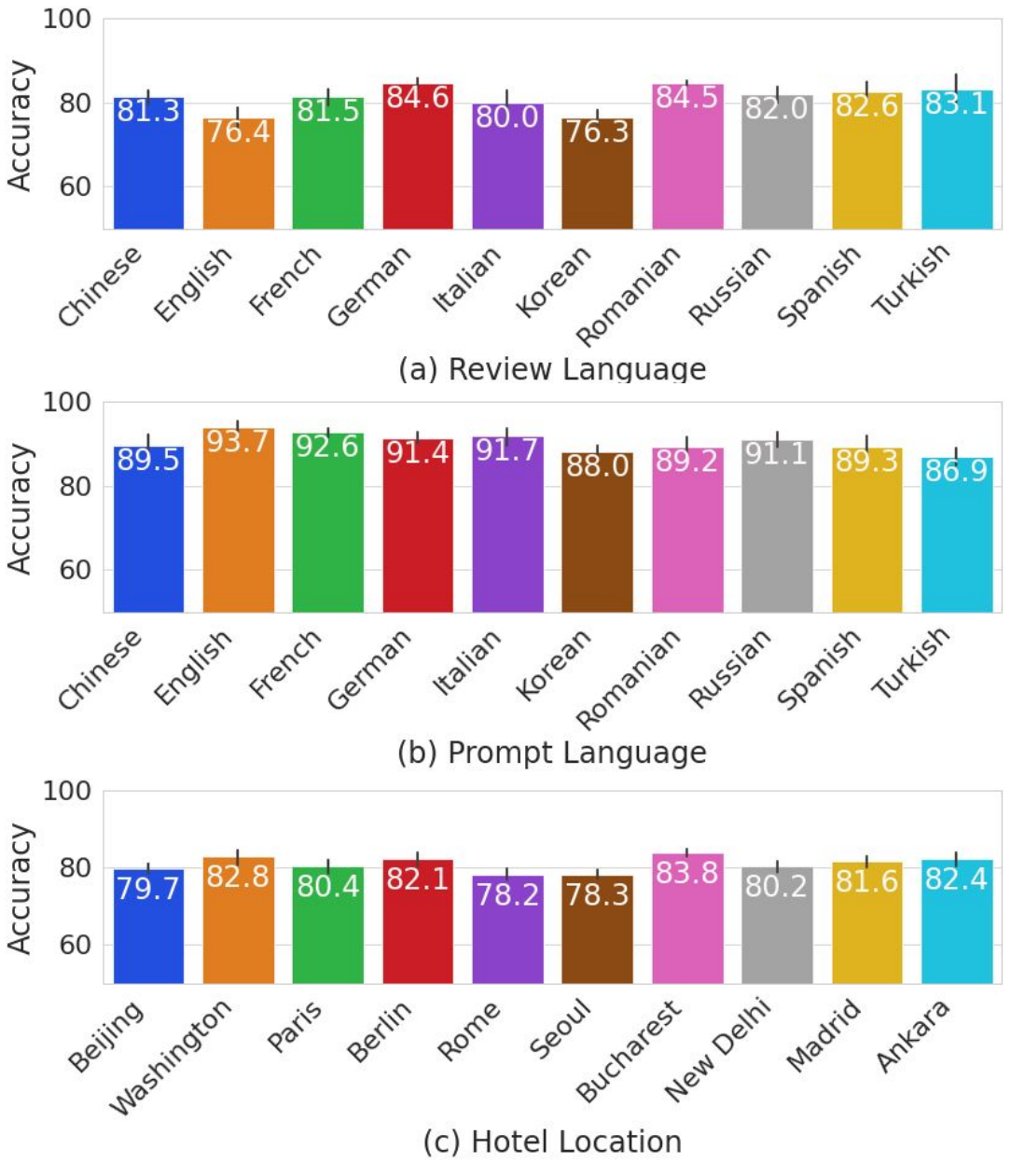}
    \caption{Accuracy with XLM-RoBERTa model per (a) review language, (b) prompt language, and (c) hotel location on the \textit{few-shot} train-test split.
    \vspace{-1.5em}}
    \label{fig:eval_loc_prompt_review}
\end{figure}

We show the XLM-RoBERTa main model \textit{few-shot} performance across sentiment, review language, prompt language, and hotel location in \Cref{fig:eval_loc_prompt_review}.
We use 10-fold cross-validation to compute confidence intervals and compute the significance using Student t-test~\citep{student1908probable} and p-value <0.05.

\vspace{-0.5em}
\paragraph{Review Language.}
As shown in \Cref{fig:eval_loc_prompt_review} (a), deception performance is lowest for \textit{Korean} and \textit{English} reviews, which implies that \textbf{GPT4 is better at generating \textit{English} and \textit{Korean} reviews}. On the other hand, deception performance is highest for \textit{German} and \textit{Romanian} reviews, indicating that \textbf{GPT4 is worse at generating \textit{German} and \textit{Romanian} reviews}. 
One possible explanation is the quantity of training data accessible for each language. Specifically, we find a moderate correlation (Pearson coefficient of 0.53) between the amount of training data and the GPT-4 performance for each language. The training data is estimated from the CommonCrawl dataset \cite{wenzek-etal-2020-ccnet}.

\vspace{-0.5em}
\paragraph{Prompt Language.}
From the perspective of a user interested in generating hotel reviews with GPT4, we measure how the language of the user prompt impacts the quality of the generated reviews. 
As shown in \Cref{fig:eval_loc_prompt_review} (b), deception performance is lowest for \textit{Turkish} and \textit{Korean}, indicating that \textbf{GPT4 is better at generating multilingual reviews when prompted in 
\textit{Turkish} and \textit{Korean}.}
On the other hand, deception detection performance is highest for \textit{English} and \textit{French} prompts, thus \textbf{GPT4 is worse at generating multilingual reviews when prompted in \textit{English} and \textit{French}.}
We measure a high negative correlation (Pearson coefficient of -0.63) between the amount of training data and and the performance of GPT-4 for each language. Therefore, understanding the impact of the language prompt on the quality of generated data is a complex issue that requires further investigation in future work.\vspace{-0.7em}

\paragraph{Hotel Location.}
When generating multilingual hotel reviews using multilingual prompts, we also measure how the location of the hotel impacts the quality of the generated reviews. As shown in \Cref{fig:eval_loc_prompt_review} (c), 
deception performance is lowest for \textit{Seoul}, \textit{Rome}, and \textit{Beijing}, indicating that \textbf{GPT4 is better at generating multilingual reviews for hotels in \textit{Seoul}, \textit{Rome}, and \textit{Beijing}.}
On the other hand, deception performance is highest for \textit{Bucharest}, \textit{Washington}, \textit{Ankara}, and \textit{Berlin}, thus \textbf{GPT4 is worse at generating multilingual reviews for hotels in \textit{Bucharest}, \textit{Washington}, \textit{Ankara}, and \textit{Berlin}.}
The results per location are moderately correlated (Pearson coefficient of 0.46) with the results per language review. Therefore, the amount of training data available for each language might also affect the model's ability to generate reviews about the locations where that language is spoken.\vspace{-0.7em}

\paragraph{Review Sentiment.} Across sentiment polarities, our main deception classification model obtains an accuracy of $79.9$ for positive reviews and $82.6$ for negative reviews. The performance difference between the two types of reviews is statistically significant.
A lower deception classification performance on positive reviews indicates that \textbf{GPT4 is more proficient in generating multilingual positive reviews than negative reviews}. This is expected given that until recently, the model did not allow negative reviews to be produced, and it also tends to be less negative than human-authored text.~\cite{markowitz2024linguistic} 

\section{Conclusion}
In this paper we focused on the identification of AI-written fake hotel reviews in multiple languages.
To facilitate research in this domain, we released \textsc{MAiDE-up}, a dataset of 10,000 real and 10,000 AI-generated fake hotel reviews balanced across ten languages.
Using this data, we performed extensive linguistic data analyses to gain insight into how AI fake hotel reviews differ from real hotel reviews. Finally, we explored the effectiveness of several models for deception detection in hotel reviews across three main dimensions: sentiment, location, and language.
Despite the difficulty humans have in distinguishing between real hotel reviews and those generated by LLMs, we discovered that these posts have noticeable differences in style, structure, and semantics and that, even with little data, fine-tuned models accurately detect deceptive reviews across different languages.
Our dataset is available for training and analyzing other models, and it can be accessed alongside our generation and classification models at
\url{https://github.com/MichiganNLP/multilingual_reviews_deception}.

\section*{Limitations}

\paragraph{Multilingual Models' Limitations. }
When analyzing data across ten different languages, we encounter significant challenges in identifying computational models and tools that can be universally applied.
In \Cref{tab:style}, we cannot find the LIWC categories required for analytic writing formula for the following languages:  \textit{German, Italian, Romanian, Korean, Russian, Turkish}. Additionally, the \texttt{textdescriptives} library does not currently support readability metrics for \textit{Chinese}, \textit{Korean}, and \textit{Turkish}.
This highlights the limitations of computational linguistic methods, which are currently predominantly English-focused. This, in turn, restricts the potential for research on multilingual data.

\paragraph{Data Generated with Closed-Source Model.}
We use GPT-4 to generate the hotel reviews, which is not an open-source LLM. We recommend future work to generate more data using open-source models like Mistral.
At the same time, we are publicly releasing the data generated with GPT-4, so that others can also build on this dataset. We chose this model due to its SOTA performance and worldwide accessibility.

\section*{Ethical Considerations}
We strongly oppose using our research findings and data to generate multilingual reviews to deceive consumers into believing they are human-authored. These unethical practices compromise the credibility of online review platforms and erode consumer trust. Instead, we advocate for transparency and authenticity in the digital marketplace. We have developed a multilingual deception detection model to combat the proliferation of fake reviews generated by bots. This model employs advanced algorithms to meticulously analyze linguistic nuances and syntactic structures, enabling the accurate differentiation between multilingual reviews created by language models and those written by human users. By providing this tool, we aim to empower businesses and online platforms to maintain ethical standards, protect consumers from deceptive practices, and foster a more trustworthy and reliable digital environment.

\bibliography{main.bib}

\appendix

\section*{Appendix}\label{appendix}
\newpage

\section{More about the Dataset}

\subsection{Real Hotel Reviews} \label{real}

\paragraph{Language}
We automatically web-crawl Booking.com for each of the ten languages by replacing the placeholder with the corresponding language abbreviation in the base URL: \specialurl{https://www.booking.com/index.{language}.html}. For example, data in Turkish can be accessed via \specialurl{https://www.booking.com/index.tr.html}. 

\begin{figure}[h!]
    \centering
    \includegraphics[width=\linewidth]{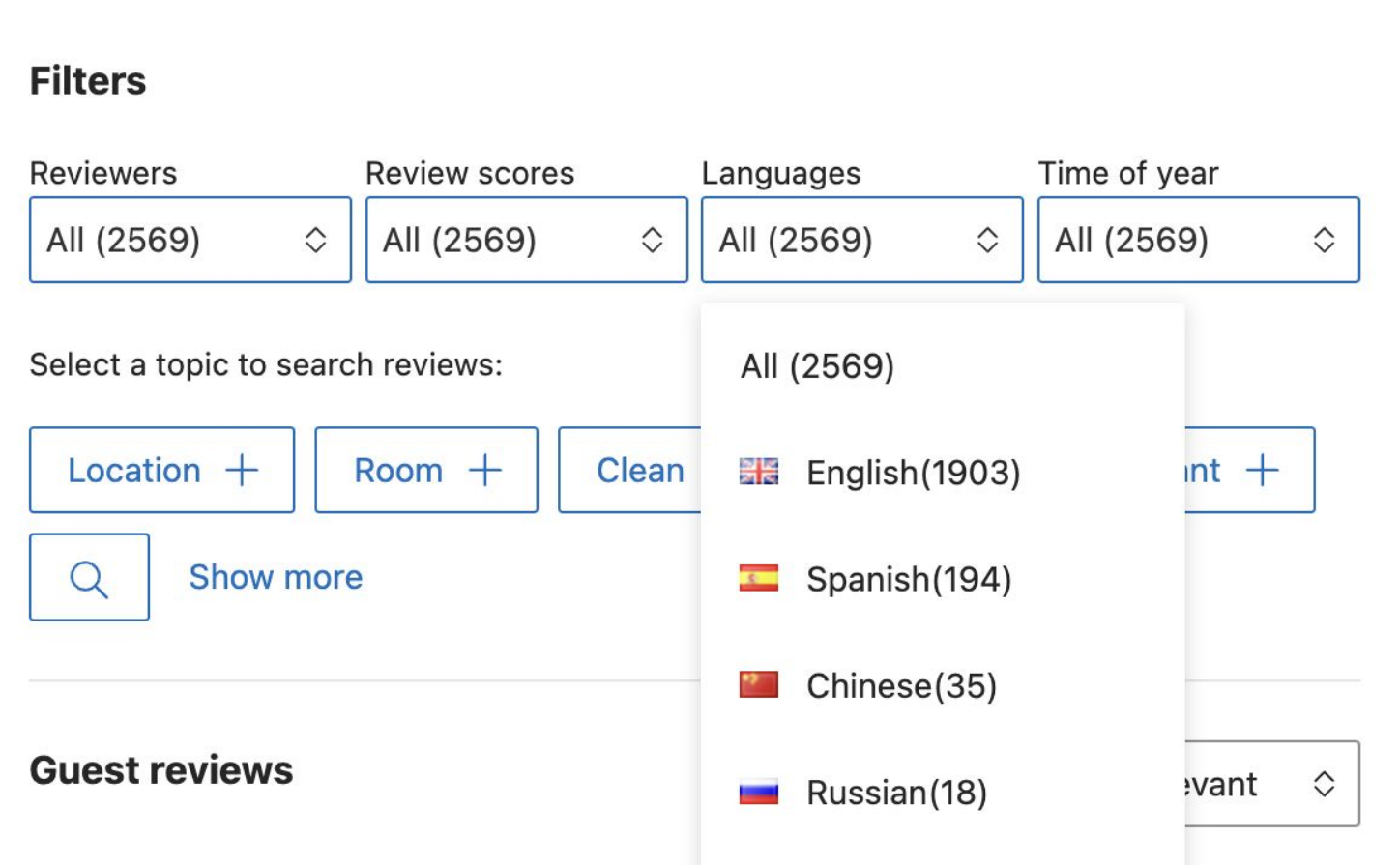}
    \caption{We use the language filter bar to select the language we specify, which is also automated by using Selenium in the data collection process.}
    \label{fig:lang-filter}
\end{figure}

To automate the web browsing process and make the data collection process more efficient, we use Selenium\footnote{\url{https://www.selenium.dev/}}. 
However, we observe that even if we browse in specific language settings, hotel reviews may still be in a mix of different languages. Therefore, we use the language filter bar to select the language we specify, which is also automated by using Selenium in the data collection process (\Cref{fig:lang-filter}).

\subsubsection{Data Quality} 

\paragraph{Automatic and manual language filtering.}
First, we filter out reviews based on length and language, that contain less than three tokens and are of a different language than the one crawled for. We use the \texttt{spaCy} library\footnote{\url{https://spacy.io/models}} to tokenize \textit{Chinese} and \textit{Korean} reviews, and \texttt{nltk}\footnote{\url{https://www.nltk.org/}} for the rest of the languages.  
Next, we automatically verify the language of each review using the \texttt{langdetect} library.\footnote{\url{https://pypi.org/project/langdetect/}} 
We find several \textit{Chinese} reviews were written in a combination of \textit{Chinese} and another language, mostly \textit{English}. 
Specifically, the ``upside'' or ``downside'' review may be in a different language, most commonly \textit{English}, as seen in Appendix \Cref{fig:mix-lang}.
We choose to keep these reviews as they reflect the most realistic \textit{Chinese} reviews written by people. Additionally, after a few manual language checks, we find that some \textit{Chinese} reviews are incorrectly classified as \textit{Korean}, and therefore chose to check all of them manually.

\paragraph{Detect fake human reviews.}
According to Booking.com, only tourists who have stayed in the hotel they booked or have visited the hotel but did not stay there can leave a review of the accommodation within three months of checking out. Additionally, Booking.com uses a combination of specialized personnel and an automated system to detect and remove fake reviews.

\begin{figure}[h!]
    \centering
    \includegraphics[width=\linewidth]{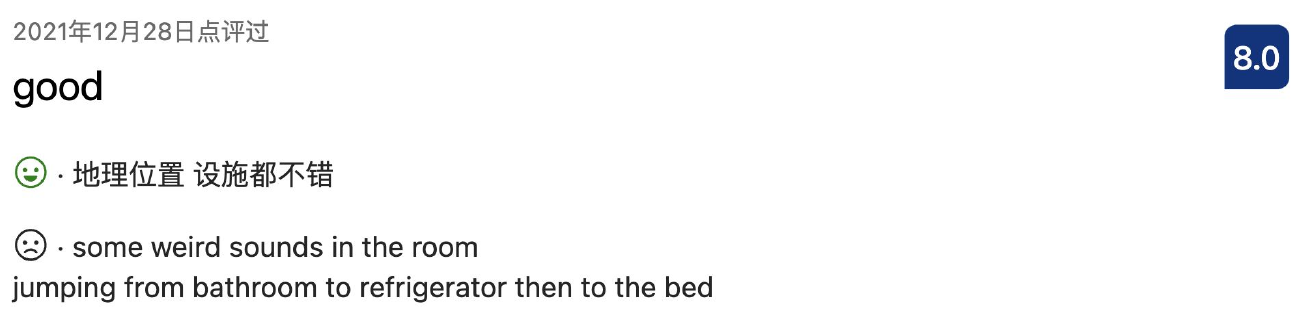}
    \caption{An example of a \textit{Chinese} review where the ``downside'' part of the review is in \textit{English}.}
    \label{fig:mix-lang}
\end{figure}

\begin{figure*}[h!]
    \centering
    \includegraphics[width=\linewidth]{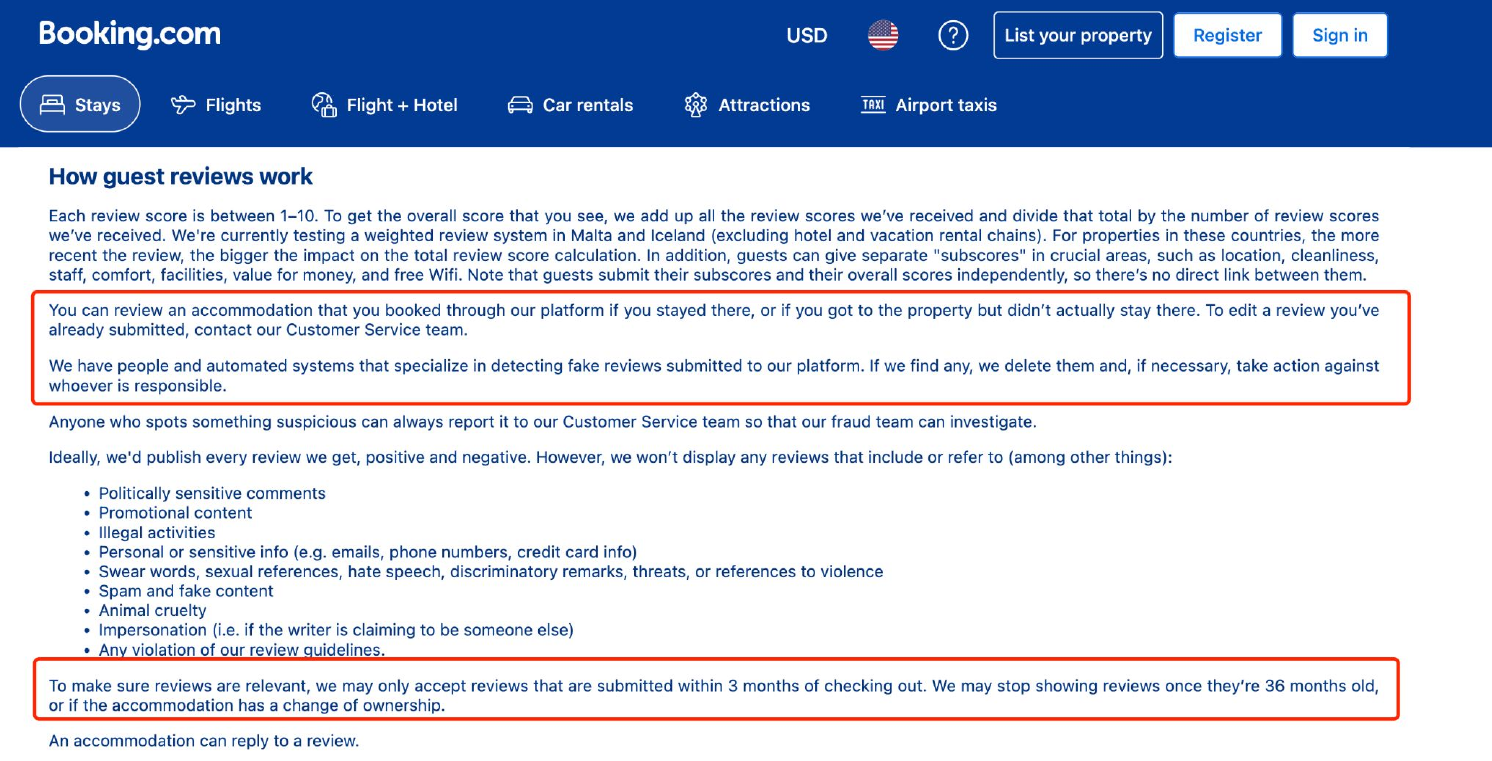}
    \caption{Specifications from Booking.com regarding checking for fake reviews}
    \label{fig:mix-lang}
\end{figure*}

\subsection{LLM-generated Hotel Reviews} \label{llm}

\paragraph{User Prompt Guidelines}

\begin{quote}
\textit{Please write 4 prompts, 2 in English and 2 in \{{language\}}, asking GPT to write a hotel review.
Each prompt should contain the following:
\begin{itemize}[noitemsep,topsep=0pt, leftmargin=*]
    \item sentiment of the review: one positive and one negative for each: English and \{{language\}})
    \item language of the review, translated in \{{language\}} for the \{{language\}} prompts
    \item city of the hotel, translated in \{{language\}} for the \{{language\}} prompts
    \item hotel name; just include it in the prompt as a placeholder: ``hotel X'', so you just translate the word hotel for the \{{language\}} prompts
\end{itemize}
}
\end{quote}

\paragraph{Challenges solved with few-shot prompting.}
The prompt design process is complex because we aim to generate reviews that capture the format of the real hotel reviews. A real review contains two sections: the ``upside'' and the ``downside'', each with corresponding sentiment. At the same time, the review's sentiment is given from the combined assessment of both the ``upside'' and the ``downside''. In addition, either the ``upside'' or ``downside'' could be null.
When we include this specification in the instruction, we find that the reviews generated with a positive sentiment predominantly contain content in the ``upside'' section, and no content in the ``downside'', and conversely for reviews with negative sentiment. 
We solve this issue by using a few-shot prompting approach, as mentioned above.

\begin{table*}[h] \footnotesize
    \begin{tabular}{c|p{14cm}}
    Lang & Prompt/ User Message \\
        \midrule
        \multirow{2}{*}{Ch} & $\color{green}\textbf{+}$ \begin{CJK*}{UTF8}{gbsn}\textit{请模仿人类用{L}为{C}的{H}酒店写一条正面评价}\end{CJK*} \\ 
        & $T:$ please write a positive review using {L} for hotel {H} in {C} that would mimic a human writing a hotel review \\ 
        & $\color{red}\textbf{-}$ \begin{CJK*}{UTF8}{gbsn}\textit{请模仿人类用{L}为{C}的{H}酒店写一条负面评价}\end{CJK*} \\ 
        & $T:$ please write a negative review using {L} for hotel {H} in {C} that would mimic a human writing a hotel review \\ 
        \midrule 
        \multirow{2}{*}{Fr} & $\color{green}\textbf{+}$ \textit{Écrivez un avis positif sur l'hôtel {H} en {C}. Veuillez écrire en {L} e parler des différents aspects de votre séjour.} \\ 
        & $T:$ Write a positive hotel review for hotel {H} in {C}. Please write it in {L} and talk about different aspects of your stay.\\ 
        & $\color{red}\textbf{-}$ \textit{Écrivez un avis négatif sur l'hôtel {H} en {C}. Veuillez écrire en {L} e parler des différents aspects de votre séjour.} \\ 
        & $T:$ Write a negative hotel review for hotel {H} in {C}. Please write it in {L} and talk about different aspects of your stay. \\ 
        \midrule 
        \multirow{2}{*}{Ge} & $\color{green}\textbf{+}$ \textit{Du bist ein Tourist, und du hast in hotel {H} in der Stadt {C} übernachtet. Das hotel hat dich richtig gefallen. Schreib ein Positives Review auf {L} über dein Stay, was fand besonderes gut and ob du das empfehlen würdest.} \\ 
        & $T:$ You are smart and helpful assistant. Your goal is to write a positive and realistic review for the hotel {H} in the language {L}, where you stayed in the city {C}. Make sure to mention why you enjoyed your stay and list all the positive features of the hotel.\\ 
        & $\color{red}\textbf{-}$ \textit{Du bist ein Tourist, und du hast in hotel {H} in der Stadt {C} übernachtet. Das hotel hat dich gar nicht gefallen. Schreib ein Negatives Review auf {L} über dein Stay, was fandest du besonderes schlecht and warum du das Hotel nicht empfehlen würdest.} \\ 
        & $T:$ You are smart and helpful assistant. Your goal is to write a negative and realistic review for the hotel {H} in the language {L}, where you stayed in the city {C}. Make sure to mention why you disliked your stay and list all the negative features of the hotel. \\ 
         \midrule
         \multirow{2}{*}{It} & $\color{green}\textbf{+}$ \textit{Scrivi una recensione positiva in {L} per l'albergo {H} di {C}. Parla di almeno tre aspetti diversi del tuo soggiorno.} \\ 
        & $T:$ Write a positive review in {L} for hotel {H} in {C}. Talk about at least three different aspects of your stay.\\ 
        & $\color{red}\textbf{-}$ \textit{Scrivi una recensione negativa in {L} per l'albergo {H} di {C}. Parla di almeno tre aspetti diversi del tuo soggiorno.} \\ 
        & $T:$ Write a negative review in {L} for hotel {H} in {C}. Talk about at least three different aspects of your stay. \\  
        \midrule 
         \multirow{2}{*}{Ko} & $\color{green}\textbf{+}$ \begin{CJK}{UTF8}{mj}\textit{지난주에 갔던 {C}의 호텔 {H}이 너무 맘에 들었다고 {L}로 리뷰좀 남겨줄래? }\end{CJK} \\ 
        & $T:$ I really enjoyed my stay at Hotel {H} in {C} last week. Can you write a review for me in {L}?\\ 
        & $\color{red}\textbf{-}$ \begin{CJK}{UTF8}{mj}\textit{{L}로 {C}에 있는 호텔 {H}이 너무 별로였다고 평점 좀 남겨줘.}\end{CJK} \\ 
        & $T:$ Can you write a review that the hotel {H} that we stayed at in {C} last week was terrible? Can you write it in {L}? \\  
         \midrule
         \multirow{2}{*}{Ro} & $\color{green}\textbf{+}$ \textit{Scrie un comentariu pozitiv in limba {L} pentru hotelul {H} din orașul {C}. Scrie la fel ca un român care a vizitat hotelul și a lăsat un comentariu.} \\ 
        & $T:$ Write a negative sentiment review in {L} language for the hotel {H} from {C}.\\ 
        & $\color{red}\textbf{-}$ \textit{Scrie un comentariu negativ in limba {L} pentru hotelul {H} din orașul {C}. Scrie la fel ca un român care a vizitat hotelul și a lăsat un comentariu.} \\ 
        & $T:$ Write a positive sentiment review in {L} language for the hotel {H} from {C}. \\
        \midrule 
        \multirow{2}{*}{Ru} & $\color{green}\textbf{+}$ \textit{\fontencoding{T2A}\selectfontПредставь что ты турист, и тебе очень понравилось твое пребывание в отеле {H} в городе {C}. Напиши реалистичный отзыв на {L} языке об этом отеле. Упомяни все черты которые тебе понравились и почему ты порекомендовала бы этот отель другим туристам.} \\ 
        & $T:$ You are a tourist and you really enjoyed staying in the hotel {H} in the city {C}. Write a simple hotel review in language {L}, where you mention all the positive features of the hotel and how much you liked them.\\ 
        & $\color{red}\textbf{-}$ \textit{ \fontencoding{T2A}\selectfontПредставь что ты турист, и тебе очень не понравился отель {H} в городе {C}. Напиши реалистичный отзыв на {L} языке об этом отеле. Упомяни все черты которые тебе не понравились и почему бы ты не порекомендовала этот отель другим туристам.} \\ 
        & $T:$ You are a tourist and you seriously dislike your stay in the hotel {H} in the city {C}. Write a simple hotel review in language {L}, where you mention all the things you disliked, and why you wouldn’t recommend this hotel to the other tourists. \\ 
        \midrule 
        \multirow{2}{*}{Sp} & $\color{green}\textbf{+}$ \textit{¿Podés escribir un comentario positivo en {L} sobre el hotel {H} de {C}?} \\ 
        & $T:$ Can you please write a positive review in {L} for the hotel {H} located in {C}? \\ 
        & $\color{red}\textbf{-}$ \textit{Escribí un comentario negativo en {L} sobre un hotel llamado {H} que está en {C}.} \\ 
        & $T:$ Write a review that's negative in sentiment in {L} language for a hotel named {H} that's located in {C}.\\ 
        \midrule 
        \multirow{2}{*}{Tu} & $\color{green}\textbf{+}$ \textit{{C} {H} için olumlu yorumu {L} olarak yaz.} \\ 
        & $T:$ Write a review in {L} about a negative experience staying in the {H} hotel in {C}. \\ 
        & $\color{red}\textbf{-}$ \textit{{C} {H} de kötü bir zaman geçirdiğini duşun. {L} olarak {H} hakkında olumsuz yorum yaz.} \\ 
        & $T:$ Imagine that you had an enjoyable stay at the hotel {H} while visiting {C}. Write a review in {L} to describe your experience to others who are considering booking a stay at {H}.\\ 
    \end{tabular}
    \caption{Prompts/ System messages in each language, together with their English translation marked with $T$.}
    \label{tab:all-prompts}
\end{table*}

\begin{table*}[h] \small
    \begin{tabular}{c|p{6cm}|p{6cm}} 
Language & Real Review & Generated Review \\
        \midrule
        \multirow{2}{*}{Chinese} & 
        $\color{green}\textbf{+}$ 
        \begin{CJK*}{UTF8}{gbsn} \textit{早餐丰富，服务态度很好，房间很整洁，离地铁站很近。优秀}\end{CJK*} & $\color{green}\textbf{+}$ \begin{CJK*}{UTF8}{gbsn}\textit{这家酒店的位置极佳，位于市中心，靠近主要的旅游景点。酒店的装修新颖，服务人员非常专业和友好。早餐也很美味，种类丰富。}\end{CJK*} \\ 
         & $\color{red}\textbf{-}$ \begin{CJK*}{UTF8}{gbsn}\textit{交通方便 价格性价比不错; 设施有些陈旧 可能因为疫情的缘故 卫生各个地方显得很陈旧的污渍 打扫看着不是那么上心 早上在酒店大堂看见任何一个服务员均很冷漠 没有打招呼的习惯希望能改进}\end{CJK*} & $\color{red}\textbf{-}$ \begin{CJK*}{UTF8}{gbsn}\textit{该酒店地理位置优越，交通便利。我订的房间缺乏维护，房间内设施感觉比较陈旧，尤其是地毯，看上去非常脏旧，且有种异味。另外，酒店早餐品种不多，食品口味偏弱。最不满的是酒店的服务态度，服务员的服务意识强度不足，没有解决问题的积极性和主动性，完全不符合四星级酒店的标准。}\end{CJK*} \\ 
        \midrule 
        \multirow{2}{*}{English} & $\color{green}\textbf{+}$ \textit{Nicely furnished room and nicely decorated lobby. Room service is affordable and the receptionists, especially Mr. Omar, are usually eager to help with any queries.} & $\color{green}\textbf{+}$ \textit{The staff was extremely helpful and accommodating. Clean and well-furnished rooms. Central location with easy access to public transport.; The breakfast offerings could be more varied.} \\ 
         & $\color{red}\textbf{-}$ \textit{The location was very central and the staff was nice and helped us.; The room smelled like cigarettes and there was mold in the bathroom} & $\color{red}\textbf{-}$ \textit{Nothing. Horrible experience.; Bad customer service, rooms were not clean and the food was below average.} \\ 
        \midrule 
        \multirow{2}{*}{French} & $\color{green}\textbf{+}$ \textit{Pas besoin de tourner pendant un moment pour trouver une place de Parking, un membre de l’hôtel se charge de prendre votre voiture; En soirée la rue n’est pas très familiale} & $\color{green}\textbf{+}$ \textit{L'hôtel est juste à côte de l'aéroport, parfait pour les vols tôt le matin. Les chambres étaient propres et confortables.; La diversité de la nourriture pourrait être améliorée.} \\ 
         & $\color{red}\textbf{-}$ \textit{Bon emplacement dans Ankara. Personnel très agréable.; Petit déjeuner moyen. Fuites dans la salle de bain. N'est pas du niveau d'un hôtel 5 étoiles.} & $\color{red}\textbf{-}$ \textit{L'emplacement est bien, situé à proximité du centre-ville d'Ankara; Le manque de propreté est notable. La nourriture n'est pas de bonne qualité et le service à la clientèle laisse à désirer. Les chambres sont bruyantes et mal isolées, et les meubles sont vieux et usés.} \\ 
        \midrule 
        \multirow{2}{*}{German} & $\color{green}\textbf{+}$ \textit{Die Lage ist super zentral! Zu Fuß nur 10 Minuten vom Kızılay Platz entfernt. Die Zimmer waren sehr sauber und sehr geschmackvoll eingerichtet. Das Frühstück war sehr reichhaltig und lecker. Das Personal ist sehr freundlich und hilfsbereit.; Keine} & $\color{green}\textbf{+}$ \textit{Das Personal war sehr freundlich und das Zimmer war sehr sauber. Die Lage ist fantastisch, nahe an vielen Sehenswürdigkeiten. Das Frühstück hatte eine gute Auswahl an Speisen.} \\ 
         & $\color{red}\textbf{-}$ \textit{Das hotel ist einfach viel zu alt. Ehrlich gesagt ist es 3 sterne; Nicht sauber habe nicht im hotel duschen können. Wir hatten gesamt zwei zimmer beide schlecht sauber gemacht worden} & $\color{red}\textbf{-}$ \textit{Die Lage des Hotels ist sehr zentral; Der Service war sehr schlecht und die Sauberkeit ließ zu wünschen übrig.} \\ 
         \midrule
         \multirow{2}{*}{Italian} & $\color{green}\textbf{+}$ \textit{Struttura bella e pulita. Vicinissimo alla metro e centro città raggiungibile a piedi. Letto e cuscini comodissimi.; Il bagno turco aveva un odore strano.. muffa?} & $\color{green}\textbf{+}$ \textit{Camera molto pulita e silenziosa, personale professionale e disponibile. La colazione era abbondante e deliziosa. Posizione eccellente nel centro della città.} \\ 
         & $\color{red}\textbf{-}$ \textit{Staff disponibile e cordiale; La camera doveva essere doppia matrimoniale, ma il letto effettivamente è una piazza e mezzo, la moquet era macchiata.} & $\color{red}\textbf{-}$ \textit{La posizione è centralissima, vicino a molti negozi e ristoranti.; Stanze poco pulite e personale maleducato. Colazione molto scarsa e Wi-Fi praticamente inesistente.} \\ 
        \midrule 
         \multirow{2}{*}{Korean} & $\color{green}\textbf{+}$ \textit{\begin{CJK}{UTF8}{mj}직원들이 자기일처럼 성실히 도와주었으며 거실,방마다있는 에어컨도 방성능이 좋았으며,조명도 밝고 시설이 깨끗해서 편안했습니다\end{CJK}; \begin{CJK}{UTF8}{mj}세탁기는 있는데 드라이어나 건조대가 없어 옷 말리기가 불편했습니다\end{CJK}} & $\color{green}\textbf{+}$ \textit{\begin{CJK}{UTF8}{mj}뉴 파크 호텔'은 매우 좋았습니다. 직원들이 친절하였고, 호텔 시설도 깨끗했습니다. 특히 아침식사가 훌륭했어요.\end{CJK}; \begin{CJK}{UTF8}{mj}방음이 잘 되지 않아 약간 소음이 있었습니다. 하지만 많이 거슬리지는 않았어요.\end{CJK}} \\ 
         & $\color{red}\textbf{-}$ \textit{\begin{CJK}{UTF8}{mj}영등포역에서 가까워요. 침구가 깨끗합니다. 방은 깨끗\end{CJK}; \begin{CJK}{UTF8}{mj}전철이 있어서 시끄럽다는 후기가 있어 씨티뷰로 신청했지만 마찬가지로 소음이 심했습니다. 실내가 좁아 불편. 주차요금을 내야합니다. 화장실이 문이 없습니다\end{CJK}} & $\color{red}\textbf{-}$ \textit{\begin{CJK}{UTF8}{mj}파리 중심부에 위치해 있어서 접근성이 좋았습니다.\end{CJK};\begin{CJK}{UTF8}{mj}객실이 너무 좁아서 실망했고, 침구가 깨끗하지 않았습니다. 또한, 냉난방 시스템이 제대로 동작하지 않아서 매우 불편했습니다.\end{CJK}} \\  
    \end{tabular}
    \caption{Random review samples with positive ($\color{green}\textbf{+}$) and negative ($\color{red}\textbf{-}$) sentiments across the 10 languages.}
    \label{tab:review-examples-all1}
\end{table*}

\begin{table*}[h] \small
    \begin{tabular}{c|p{6cm}|p{6cm}} 
Language & Real Review & Generated Review \\
        \midrule
         \multirow{4}{*}{Romanian} & $\color{green}\textbf{+}$ \textit{Gazde foarte amabile, primitoare, camera mare, curată, paturi confortabile, parcare proprie în fața hotelului, micul dejun, bufet, diversificat.} & $\color{green}\textbf{+}$ \textit{Servicii excelente, camere curate și spațioase. Spa-ul și piscina sunt minunate pentru relaxare."; Nu am avut nicio problemă sau dezamăgire.} \\ 
         & $\color{red}\textbf{-}$ \textit{E ffff cald in camere! Și aparatele stricate, camerele mici!!} & $\color{red}\textbf{-}$ \textit{Amplasarea hotelului este convenabilă, relativ aproape de majoritatea obiectivelor turistice.; Confortul camerei a fost sub orice critica, patul a fost foarte inconfortabil, iar curăţenia a fost precară. De asemenea, personalul a fost nepoliticos si neajutorător.} \\
        \midrule 
        \multirow{2}{*}{Russian} & $\color{green}\textbf{+}$ \textit{\fontencoding{T2A}\selectfont Шикарный завтрак, удобное расположение, доброжелательный персонал, согласовали ранний заезд; Немного грязное окно и тусклое освещение, но это мелочи} & $\color{green}\textbf{+}$ \textit{\fontencoding{T2A}\selectfontОтель с удобным расположением, близко к основным достопримечательностям. Очень чистые и комфортные номера. Отзывчивый и вежливый персонал.; Сложно было найти русскоговорящий персонал. Завтраки могли бы быть более разнообразными.} \\ 
         & $\color{red}\textbf{-}$ \textit{\fontencoding{T2A}\selectfontметро рядом , санлитунь рядом.; персонал за стойками регистрации, это просто треш, мало того, что цены в чеке не совпадали с ценой на букинге, так еще и чекаут оформляли 50 минут! это дно товарищи!} & $\color{red}\textbf{-}$ \textit{\fontencoding{T2A}\selectfontНет положительных сторон.; Большая часть персонала не говорит по-английски, что особенно усложнило процесс общения. На завтраке был ограниченный выбор блюд, а в номерах отличался сильный запах курения.} \\ 
        \midrule 
        \multirow{2}{*}{Spanish} & $\color{green}\textbf{+}$ \textit{Me gustó todo, super bonito, limpio, cómodo y en una zona muy animada, con un desayuno buffet genial; Nada} & $\color{green}\textbf{+}$ \textit{Un hotel muy bueno. Estaba muy limpio y moderno. En el corazón de la ciudad, cerca de las tiendas y restaurantes. El personal fue amigable y la comida era buena.; La única queja que tenía era que el wifi en mi habitación era un poco lento.} \\ 
         & $\color{red}\textbf{-}$ \textit{Hotel con bien trato, y muy céntrico, limpio y comodo; Discoteca justo debajo de la habitación muchísimo ruido y música disco mucho volumen hasta altas horas. No se puede descansar} & $\color{red}\textbf{-}$ \textit{Buena ubicación central y la habitación estaba limpia; El desayuno era muy básico y la presión del agua en la ducha era bastante baja} \\ 
        \midrule 
        \multirow{2}{*}{Turkish} & $\color{green}\textbf{+}$ \textit{Kahvaltısı çok güzeldi. Tam bir Fransız kahvaltısıydı. Personel çok güler yüzlüydü. Otel çok merkezi bir konumdaydı. Tekrar Paris'e gelirsem tercihim yine buradan yana olur.} & $\color{green}\textbf{+}$ \textit{Otelin konumu ve erişilebilirliği mükemmel. Odalar temiz, konforlu ve fonksiyonel.; Oda sıcaklık ayarları biraz daha iyi olabilirdi.} \\ 
         & $\color{red}\textbf{-}$ \textit{Bookingde sigarasız yazmasına rağmen ortak havalandırmadan sigara dumanı geliyordu wi-fi çalışmıyordu} & $\color{red}\textbf{-}$ \textit{Otelin konumu iyi.; Oda temizliği yetersizdi, yemekler çok tuzluydu ve personel pek yardımcı olmadı. Bu nedenlerden dolayı diğer turistlere bu oteli önermiyorum.} \\ 
    \end{tabular}
    \caption{Random review samples with positive ($\color{green}\textbf{+}$) and negative ($\color{red}\textbf{-}$) sentiments across the 10 languages.}
    \label{tab:review-examples-all2}
\end{table*}

\begin{table*}[h!]
\resizebox{\textwidth}{!}{%
\begin{tabular}{|c|cc|cc|cc}
\toprule
& \multicolumn{2}{c|}{\textbf{uni-gram}} 
& \multicolumn{2}{c|}{\textbf{bi-gram}}
& \multicolumn{2}{c|}{\textbf{tri-gram}}
\\
\midrule
\multirow{1}{*}{\textbf{Lang}} &
  \multicolumn{1}{c|}{\textbf{Real}} &
  \multicolumn{1}{c|}{\textbf{Gen}} &
  \multicolumn{1}{c|}{\textbf{Real}} &
  \multicolumn{1}{c|}{\textbf{Gen}} &
  \multicolumn{1}{c|}{\textbf{Real}} &
  \multicolumn{1}{c|}{\textbf{Gen}} \\
\midrule
\worldflag[length=0.5cm, width=0.3cm]{CN} & \multicolumn{1}{c|}{\shortstack{\begin{CJK*}{UTF8}{gbsn}酒店\end{CJK*}\\ \begin{CJK*}{UTF8}{gbsn}房间\end{CJK*}\\ \begin{CJK*}{UTF8}{gbsn}位置\end{CJK*}}} & 
\multicolumn{1}{c|}{\shortstack{\begin{CJK*}{UTF8}{gbsn}酒店\end{CJK*}\\ \begin{CJK*}{UTF8}{gbsn}位置\end{CJK*}\\ \begin{CJK*}{UTF8}{gbsn}服务\end{CJK*}}} & 
\multicolumn{1}{c|}{\shortstack{\begin{CJK*}{UTF8}{gbsn}(地理, 位置)\end{CJK*}\\ \begin{CJK*}{UTF8}{gbsn}(工作, 人员)\end{CJK*}\\ \begin{CJK*}{UTF8}{gbsn}(位置, 不错)\end{CJK*}}} & 
\multicolumn{1}{c|}{\shortstack{\begin{CJK*}{UTF8}{gbsn}(地理, 位置)\end{CJK*}\\ \begin{CJK*}{UTF8}{gbsn}(酒店, 位置)\end{CJK*}\\ \begin{CJK*}{UTF8}{gbsn}(服务, 人员)\end{CJK*}}} & 
\multicolumn{1}{c|}{\shortstack{\begin{CJK*}{UTF8}{gbsn}(地理, 位置, 不错)\end{CJK*}\\ \begin{CJK*}{UTF8}{gbsn}(地理, 位置, 优越)\end{CJK*}\\\begin{CJK*}{UTF8}{gbsn}(员工, 服务, 态度)\end{CJK*}}} & 
\multicolumn{1}{c|}{\shortstack{\begin{CJK*}{UTF8}{gbsn}(酒店, 地理, 位置)\end{CJK*}\\ \begin{CJK*}{UTF8}{gbsn}(服务, 人员, 态度)\end{CJK*}\\ \begin{CJK*}{UTF8}{gbsn}(地理, 位置, 优越)\end{CJK*}}}
\\ 
\midrule
\worldflag[length=0.5cm, width=0.3cm]{US} & \multicolumn{1}{c|}{\shortstack{room\\ location\\ hotel}} & \multicolumn{1}{c|}{\shortstack{room\\ hotel\\ location}} & \multicolumn{1}{c|}{\shortstack{(great, location)\\ (room, clean)\\ (location, good)}} & \multicolumn{1}{c|}{\shortstack{(room, clean)\\(staff, friendly)\\(customer, service)}} 
& \multicolumn{1}{c|}{\shortstack{(staff, friendly, helpful)\\ (staff, nice, helpful)\\ (staff, speak, english)}}
& \multicolumn{1}{c|}{\shortstack{(room, clean, comfortable)\\(leave, lot, desire)\\(staff, friendly, helpful)}}
\\ 
\midrule
\worldflag[length=0.5cm, width=0.3cm]{FR} & \multicolumn{1}{c|}{\shortstack{chambre\\ petit\\ hôtel}} & \multicolumn{1}{c|}{\shortstack{chambre\\ hôtel\\ personnel}} & \multicolumn{1}{c|}{\shortstack{(petit, déjeuner)\\ (salle, bain)\\ (bien, situer)}} & \multicolumn{1}{c|}{\shortstack{(petit, déjeuner)\\(chambre, propre)\\(chambre, spacieux)}} 
& \multicolumn{1}{c|}{\shortstack{(rapport, qualité, prix)\\ (hôtel, bien, situer)\\ (bon, petit, déjeuner)}}
& \multicolumn{1}{c|}{\shortstack{(petit, déjeuner, varier)\\(chambre, propre, confortable)\\(chambre, spacieux, propre)}}
\\ 
\midrule
\worldflag[length=0.5cm, width=0.3cm]{DE} & \multicolumn{1}{c|}{\shortstack{Zimmer\\ Lage\\ Hotel}} & \multicolumn{1}{c|}{\shortstack{Hotel\\ Zimmer\\ Lage}} & \multicolumn{1}{c|}{\shortstack{(freundlich, Personal)\\ (Personal, freundlich)\\ (zentral, Lage)}} & \multicolumn{1}{c|}{\shortstack{(Lage, Hotel)\\(freundlich, hilfsbereit)\\(Personal, freundlich)}} 
& \multicolumn{1}{c|}{\shortstack{(Personal, freundlich, hilfsbereit)\\ (Lage, freundlich, Personal)\\ (Personal, super, freundlich)}}
& \multicolumn{1}{c|}{\shortstack{(Personal, freundlich, hilfsbereit)\\(lassen, wünschen, übrig)\\(Lage, Hotel, zentral)}}
\\ 
\midrule
\worldflag[length=0.5cm, width=0.3cm]{IT} & \multicolumn{1}{c|}{\shortstack{posizione\\ camera\\ colazione}} & \multicolumn{1}{c|}{\shortstack{posizione \\ hotel \\ camera}} & \multicolumn{1}{c|}{\shortstack{(ottimo, posizione)\\ (posizione, ottimo)\\ (personale, gentile)}} & \multicolumn{1}{c|}{\shortstack{(camera, pulito)\\(posizione, centrale)\\(posizione, hotel)}} 
& \multicolumn{1}{c|}{\shortstack{(rapporto, qualità, prezzo)\\ (personale, gentile, disponibile)\\ (ottimo, rapporto, qualità)}}
& \multicolumn{1}{c|}{\shortstack{(personale, cordiale, disponibile)\\(camera, pulito, confortevole)\\(personale, gentile, disponibile)}}
\\ 
\midrule
\worldflag[length=0.5cm, width=0.3cm]{KR} & 
\multicolumn{1}{c|}{\shortstack{\begin{CJK}{UTF8}{mj}너무\end{CJK}\\ \begin{CJK}{UTF8}{mj}좋았습니다\end{CJK}\\ \begin{CJK}{UTF8}{mj}위치가\end{CJK}}} &
\multicolumn{1}{c|}{\shortstack{\begin{CJK}{UTF8}{mj}매우\end{CJK}\\ \begin{CJK}{UTF8}{mj}또한\end{CJK}\\ \begin{CJK}{UTF8}{mj}호텔\end{CJK}}} & 
\multicolumn{1}{c|}{\shortstack{\begin{CJK}{UTF8}{mj}(바로, 앞에)\end{CJK}\\ \begin{CJK}{UTF8}{mj}(너무, 좋았어요)\end{CJK}\\ \begin{CJK}{UTF8}{mj}(위치가, 너무)\end{CJK}}} & 
\multicolumn{1}{c|}{\shortstack{\begin{CJK}{UTF8}{mj}(호텔의, 위치는)\end{CJK}\\\begin{CJK}{UTF8}{mj}(방음이, 되지)\end{CJK}\\\begin{CJK}{UTF8}{mj}(또한, 직원들의)\end{CJK}}} 
& \multicolumn{1}{c|}{\shortstack{\begin{CJK}{UTF8}{mj}(위치가, 너무, 좋았고)\end{CJK}\\ \begin{CJK}{UTF8}{mj}(바로, 앞에, 있어서)\end{CJK}\\ \begin{CJK}{UTF8}{mj}(좋음, 직원, 친절함)\end{CJK}}}
& \multicolumn{1}{c|}{\shortstack{\begin{CJK}{UTF8}{mj}(방음이, 되지, 않아)\end{CJK}\\ \begin{CJK}{UTF8}{mj}(방음이, 되지, 않아서)\end{CJK}\\ \begin{CJK}{UTF8}{mj}(친절하고, 도움이, 되었습니다)\end{CJK}}}
\\ 
\midrule
\worldflag[length=0.5cm, width=0.3cm]{RO} & \multicolumn{1}{c|}{\shortstack{cameră\\ mic\\ hotel}} & \multicolumn{1}{c|}{\shortstack{cameră\\ personal\\ hotel}} & \multicolumn{1}{c|}{\shortstack{(mic, dejun)\\ (personal, amabil)\\ (cameră, mic)}} & \multicolumn{1}{c|}{\shortstack{(mic, dejun)\\(personal, amabil)\\(personal, prietenos)}} 
& \multicolumn{1}{c|}{\shortstack{(mic, dejun, bun)\\ (mic, dejun, bogat)\\ (mic, dejun, ok)}}
& \multicolumn{1}{c|}{\shortstack{(cameră, curat, confortabil)\\(aproape, centru, oraș)\\(mic, dejun, putea)}}
\\ 
\midrule
\worldflag[length=0.5cm, width=0.3cm]{RU} & \multicolumn{1}{c|}{\shortstack{\fontencoding{T2A}\selectfont номер\\ \fontencoding{T2A}\selectfont отель\\ \fontencoding{T2A}\selectfont завтрак}} &
\multicolumn{1}{c|}{\shortstack{\fontencoding{T2A}\selectfont отель\\ \fontencoding{T2A}\selectfont номер\\ \fontencoding{T2A}\selectfont персонал}} & 
\multicolumn{1}{c|}{\shortstack{\fontencoding{T2A}\selectfont (постельный, бельё)\\ \fontencoding{T2A}\selectfont (горячий, вода)\\ \fontencoding{T2A}\selectfont (приветливый, персонал)}} & 
\multicolumn{1}{c|}{\shortstack{\fontencoding{T2A}\selectfont (центр, город)\\ \fontencoding{T2A}\selectfont (желать, хороший)\\(wi, fi)}} 
& \multicolumn{1}{c|}{\shortstack{\fontencoding{T2A}\selectfont(соотношение, цена, качество)\\ \fontencoding{T2A}\selectfont (постельный, бельё, полотенце)\\\fontencoding{T2A}\selectfont (персонал, говорить, английский)}}
& \multicolumn{1}{c|}{\shortstack{\fontencoding{T2A}\selectfont(оставлять, желать, хороший)\\ \fontencoding{T2A}\selectfont(дружелюбный, готовый, помочь)\\ \fontencoding{T2A}\selectfont(бесплатный, wi, fi)}}
\\ 
\midrule
\worldflag[length=0.5cm, width=0.3cm]{ES} & \multicolumn{1}{c|}{\shortstack{habitación\\ ubicación\\ hotel}} & \multicolumn{1}{c|}{\shortstack{habitación\\ hotel\\ ubicación}} & \multicolumn{1}{c|}{\shortstack{(personal, amable)\\ (habitación, pequeño)\\ (aire, acondicionado)}} & \multicolumn{1}{c|}{\shortstack{(personal, amable)\\(dispuesto, ayudar)\\(ubicación, hotel)}} 
& \multicolumn{1}{c|}{\shortstack{(relación, calidad, precio)\\ (personal, amable, habitación)\\ (cerca, torre, eiffel)}}
& \multicolumn{1}{c|}{\shortstack{(amable, dispuesto, ayudar)\\(personal, amable, dispuesto)\\(personal, amable, servicial)}}
\\ 
\midrule
\worldflag[length=0.5cm, width=0.3cm]{TR} & \multicolumn{1}{c|}{\shortstack{bir\\ konum\\ od}} & \multicolumn{1}{c|}{\shortstack{oda\\ otel\\ olmak}} & \multicolumn{1}{c|}{\shortstack{(konum, iyi)\\ (gülmek, yüz)\\ (od, Küçük)}} & \multicolumn{1}{c|}{\shortstack{(otel, konum)\\ (oda, temiz)\\ (yardımcı, olmak)}} 
& \multicolumn{1}{c|}{\shortstack{(çalışmak, gülmek, yüz)\\ (personel, gülmek, yüz)\\ (yer, yürümek, mesafe)}}
& \multicolumn{1}{c|}{\shortstack{(personel, son, derece)\\ (oda, te, konfor)\\ (oda, temiz, yeter)}}
\\ 
\bottomrule
\end{tabular}%
}
\caption{Most frequent n-grams for real and generated data, for each review language. Country flag according to review location.}
\label{tab:n-grams}
\end{table*}

\section{Data Analysis}

\paragraph{Word frequency.}
We compute the most frequent n-grams across each language after pre-processing the text, which is shown in \Cref{tab:n-grams}.

\paragraph{Topic Modeling}\label{topic}

\begin{figure*}
\centering
\includegraphics[width=\linewidth]{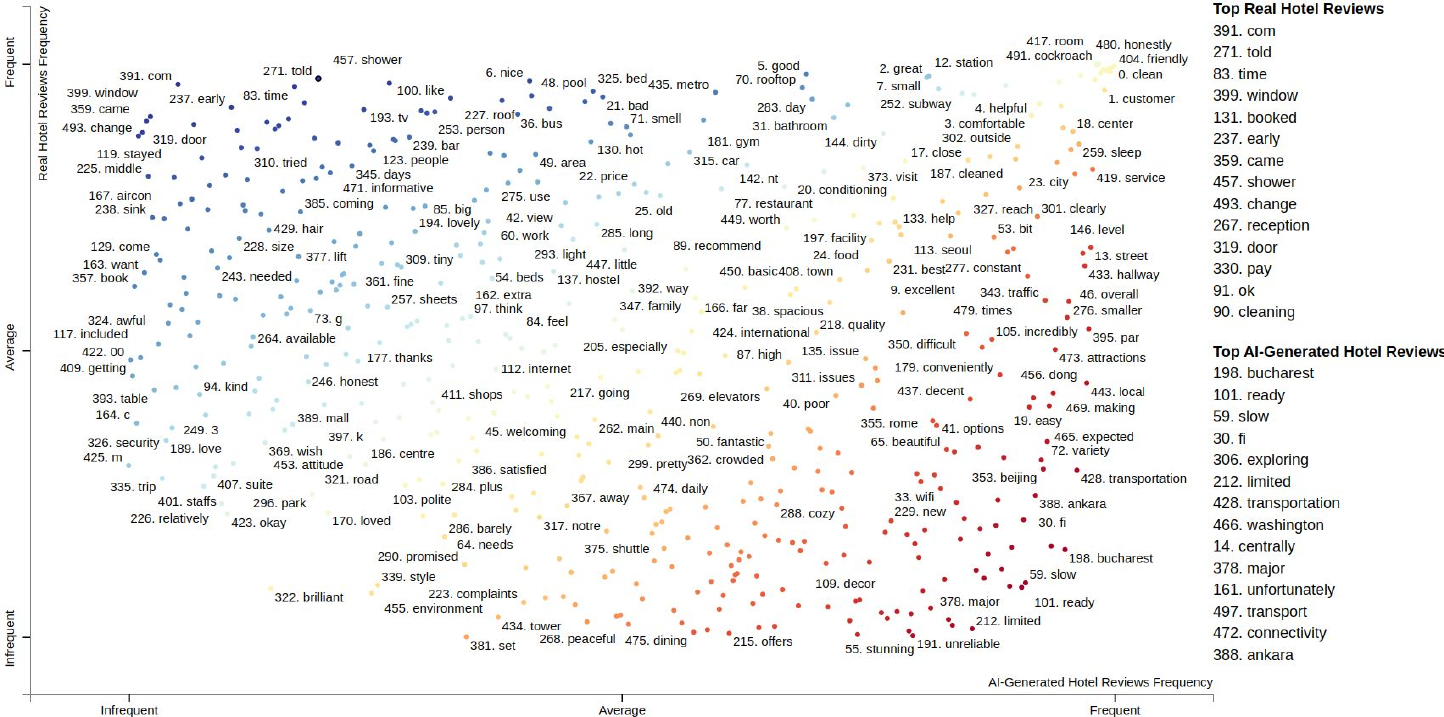}
\caption{Visualization of \textbf{topics} used in the real and LLM-generated English reviews. Points are colored red or blue based on the association of their corresponding topics with AI-generated or real hotel reviews. The most associated topics are listed under \textbf{Top AI-Generated} and \textbf{Top Real} headings. Interactive version: \ttfamily\small\url{https://anonymous.4open.science/r/hotel_reviews_deception}.}
\label{fig:topic_english}
\end{figure*}

\begin{figure*}
\centering
\includegraphics[width=\linewidth]{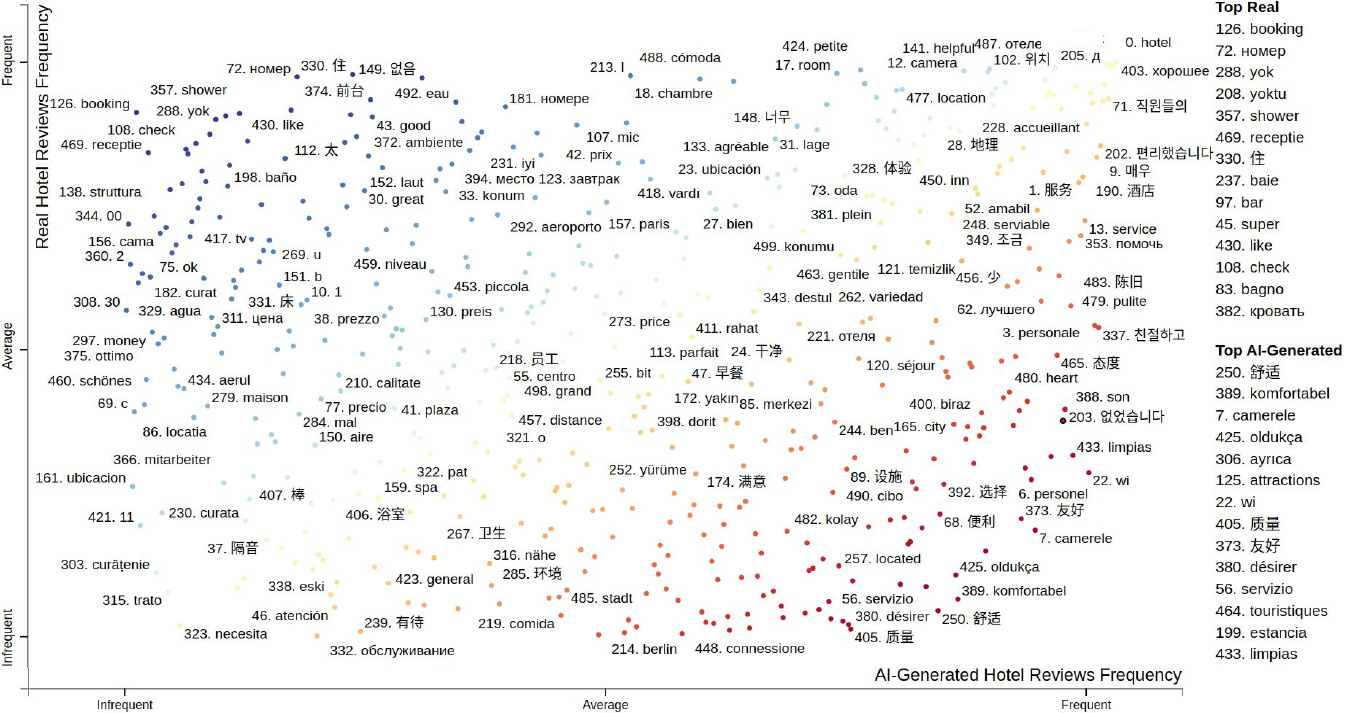}
\caption{Visualization of \textbf{topics} used in the real and LLM-generated reviews across all languages. Points are colored red or blue based on the association of their corresponding topics with AI-generated or real hotel reviews. The most associated topics are listed under \textbf{Top AI-Generated} and \textbf{Top Real} headings. Interactive version: \ttfamily\small\url{https://anonymous.4open.science/r/hotel_reviews_deception}.}
\label{fig:topic_multi}
\end{figure*}

\end{document}